\documentclass{article}
\usepackage{te_preprint}

\usepackage{amsmath,amsfonts,bm}
\def\1{\bm{1}}

\usepackage{hyperref}
\hypersetup{pdftitle={TokenEconomizer: Experience-Driven Black-Box Cost Control for Coding Agents},
  pdfauthor={Aaron J. Li, Weijia Xu, Chandan Singh, Zixiao Chen, Jeevana Priya Inala, Charlotte Siska,
    Acar Tamersoy, Bin Yu, Ion Stoica, Michel Galley, Jianfeng Gao}}
\usepackage{url}
\usepackage{graphicx}
\usepackage{booktabs}
\usepackage{xcolor}
\usepackage{microtype}
\usepackage{xspace}
\usepackage{subcaption}
\usepackage[capitalize]{cleveref}
\crefname{section}{Sec.}{Secs.}
\Crefname{section}{Sec.}{Secs.}
\crefname{subsection}{Sec.}{Secs.}
\Crefname{subsection}{Sec.}{Secs.}
\crefname{appendix}{Appendix}{Appendices}
\Crefname{appendix}{Appendix}{Appendices}
\crefname{subappendix}{Appendix}{Appendices}
\Crefname{subappendix}{Appendix}{Appendices}
\crefname{table}{Table}{Tables}
\Crefname{table}{Table}{Tables}
\crefname{figure}{Fig.}{Figs.}
\Crefname{figure}{Fig.}{Figs.}
\crefname{equation}{Eq.}{Eqs.}
\Crefname{equation}{Eq.}{Eqs.}

\creflabelformat{equation}{#2(#1)#3}
\usepackage{enumitem}
\usepackage{listings}
\usepackage{tabularx}

\newcommand{\method}{\textsc{TokenEconomizer}\xspace}
\newcommand{\cost}{C_{\mathrm{eff}}}

\definecolor{teblue}{HTML}{3366A8}
\definecolor{telight}{HTML}{8FB4DC}
\definecolor{defgrey}{HTML}{6E767E}
\definecolor{promptamber}{HTML}{DF9843}
\definecolor{slotamber}{HTML}{A8662A}
\lstdefinestyle{skillblock}{basicstyle=\ttfamily\scriptsize,frame=l,framerule=1.6pt,
  framesep=4pt,xleftmargin=5.6pt,xrightmargin=3pt,framextopmargin=1.5pt,framexbottommargin=1.5pt,
  framexrightmargin=3pt,aboveskip=2pt,belowskip=0pt,upquote=true,
  columns=fullflexible,keepspaces=true,breaklines=true,breakatwhitespace=true,breakindent=2em}
\lstnewenvironment{SkillText}{\lstset{style=skillblock,breakindent=0pt,
  rulecolor=\color{teblue},backgroundcolor=\color{teblue!8}}}{}
\lstnewenvironment{Trace}{\lstset{style=skillblock,rulecolor=\color{defgrey!60},
  backgroundcolor=\color{defgrey!7},moredelim=[is][\color{black!55}]{@@}{@@}}}{}
\lstnewenvironment{TraceTE}{\lstset{style=skillblock,rulecolor=\color{telight},
  backgroundcolor=\color{teblue!3},moredelim=[is][\color{black!55}]{@@}{@@}}}{}
\lstnewenvironment{PromptText}{\lstset{style=skillblock,breakindent=0pt,
  rulecolor=\color{promptamber},backgroundcolor=\color{promptamber!7},
  moredelim=[is][\color{slotamber}]{@@}{@@}}}{}
\newcommand{\blocklabel}[1]{\par\smallskip\noindent{\scriptsize #1}\par}

\title{\method: Experience-Driven Black-Box Cost Control for Coding Agents}

\author{%
\authorname{Aaron J. Li}{1,*}\authorsep
\authorname{Weijia Xu}{2}\authorsep
\authorname{Chandan Singh}{2}\authorsep
\authorname{Zixiao Chen}{3}\authorsep
\authorname{Jeevana Priya Inala}{2}\authorsep
\authorname{Charlotte Siska}{3}\authorsep
\authorname{Acar Tamersoy}{3}\authorsep
\authorname{Bin Yu}{1}\authorsep
\authorname{Ion Stoica}{1}\authorsep
\authorname{Michel Galley}{2}\authorsep
\authorname{Jianfeng Gao}{2}}
\affiliations{\textsuperscript{1}UC Berkeley\qquad \textsuperscript{2}Microsoft Research\qquad
  \textsuperscript{3}Microsoft Security}
\links{\weblink{Project page}{https://tokeneconomizer.github.io}\\
  \weblink{Code}{https://github.com/microsoft/TokenEconomizer}}
\logos{\logo{21pt}{figures/logo_berkeley}}{\logo{14pt}{figures/logo_microsoft}}
\titlenote{Work done during an internship at Microsoft Research. Correspondence to:
  \href{mailto:aaronjli@berkeley.edu}{\texttt{aaronjli@berkeley.edu}}.}

\begin{document}

\maketitle

\begin{abstract}
Advances in frontier coding agents have sharply increased the token costs of AI-assisted software engineering. Their token efficiency is typically evaluated on isolated tasks, whereas real-world developers use them to continuously build, update, and maintain large-scale projects. This mismatch motivates us to evaluate agents over sequences of realistic coding tasks, where we find that agents exhibit recurring environment-dependent token costs that existing methods struggle to reduce. To minimize these costs, we introduce \method, an external controller that cuts environment-specific token costs over a sequence of tasks. These cuts come from determining (i) which operational micro-skills to inject, (ii) which persistently irrelevant tools to prune, and (iii) when to intervene in unproductive failure loops. Across Claude Code, Codex, and GitHub Copilot CLI paired with frontier models, \method reduces the average token cost by 18.4–30.9\% with comparable task success. Savings hold in every repository under every tested agent–model setting and generalize to diverse domains such as sequential web development and security auditing, where costs fall by 6.4–21.2\% depending on task structure. Our work establishes adaptive black-box cost control as a promising direction for reducing token costs of coding agents in realistic settings.

\end{abstract}

\section{Introduction}

\begin{figure}[t]
\centering
\begin{subfigure}[b]{0.518\linewidth}
  \centering
  \includegraphics[width=\linewidth]{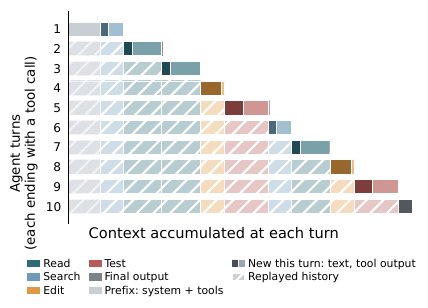}
  \caption{Anatomy of a billed trajectory (schematic)}
  \label{fig:pilot-profile-a}
\end{subfigure}\hfill
\begin{subfigure}[b]{0.464\linewidth}
  \centering
  \includegraphics[width=\linewidth]{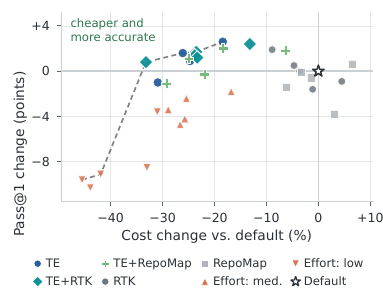}
  \caption{Cost--quality tradeoff on SWE-Bench}
  \label{fig:main-frontier}
\end{subfigure}
\caption{Trajectory anatomy and the cost--quality tradeoff. (a) Schematic of how the persistent prefix and interaction history accumulate across agent turns. (b) Cost and Pass@1 change of each method relative to high-effort default execution, one point per method in each of the five agent--model settings; the dashed line marks the Pareto frontier.}
\label{fig:pilot-profile}
\end{figure}

Frontier large language models (LLMs), coupled with agent harnesses that support tool use~\citep{yang2024sweagent,wang2024executable,wang2025openhands}, have become increasingly capable of completing realistic, long-horizon software engineering tasks~\citep{jimenez2024swebench,zhou2024webarena,xu2025webbench,deng2025swebenchpro,merrill2026terminal,huang2026deepswe}. These components are commonly packaged as commercial products, including \href{https://code.claude.com/docs/en/overview}{Claude Code}, \href{https://github.com/openai/codex}{Codex}, and \href{https://docs.github.com/en/copilot/concepts/agents/copilot-cli}{GitHub Copilot CLI}. These agents are designed primarily for generic task success, but their increased capability comes at a substantial token cost. As shown in \cref{fig:pilot-profile-a}, the billed context at each turn includes persistent system and user instructions together with an expanding history of model outputs and tool interactions, causing earlier content to be charged repeatedly across the trajectory.

Token efficiency has primarily been studied and optimized at the level of individual tasks. Model-level controls such as lower reasoning effort save tokens at a cost in task success (\cref{tab:main}), while harness-level controls such as tool availability and auxiliary instructions typically require manual configuration. More targeted approaches to trajectory compression, observation filtering, and specialized repository exploration often require agent-loop modifications or additional trained components~\citep{xiao2025agentdiet,kovacs2026squeez,li2026fastcode}, limiting their plug-and-play deployment. More importantly, task-local optimization overlooks the continuity of software development, where successive changes occur within an evolving codebase~\citep{xu2025webbench,joshi2025swebench-cl,zhu2026swecontext,yan2026agentstream}. Such continuity creates an opportunity for cost control: within a shared environment, such as a code repository, completed executions can provide reusable experience for future tasks. We therefore optimize cumulative cost across a sequence of tasks in that environment, rather than treating each execution in isolation.

We first test whether fixed, externally applied cost-control strategies can deliver consistent token savings. Our cost profiling covers 60 SWE-Bench tasks across 6 repositories~\citep{jimenez2024swebench} and evaluates three representative fixed strategies from popular open-source plugins: command-output compression with \href{https://github.com/rtk-ai/rtk}{Rust Token Killer} (RTK), semantic code navigation with \href{https://github.com/oraios/serena}{Serena}, and structural context injection with \href{https://aider.chat/docs/repomap.html}{Aider's repository map}. Empirically, we find that no fixed strategy reduces cost consistently: each has both positive and negative repository-level effects, and their pooled effects are small (\cref{fig:pilot-strategies}). Meanwhile, the default traces reveal three recurring sources of headroom: operational knowledge is repeatedly rediscovered, most loaded tools remain unused, and failed trajectories consume disproportionate cost. Taken together, these findings point to strategies that adapt to each environment from its accumulated traces, rather than fixed strategies applied uniformly.

Motivated by these findings, we introduce \method, an experience-driven external controller for coding agents. It treats each agent as a black box: it acts through exposed prompt and tool settings, monitors execution events, and does not modify model weights or agent-loop code. At its core, a controller applies general cost-control meta-strategies to continually refine environment-specific policies from completed execution traces. These policies adaptively determine (i) which one-line operational facts about the environment to activate as micro-skills, (ii) which loaded but irrelevant tools to prune, and (iii) when to intervene in unproductive failure loops. We evaluate \method with Claude Code, Codex, and GitHub Copilot CLI, paired with frontier models, on 277 SWE-Bench tasks across 6 repositories, 102 Web-Bench tasks across 5 web-development projects, and 200 tasks from two security-auditing benchmarks. Including all controller overhead, our method reduces mean per-task cost on SWE-Bench by 18.4--30.9\%, while improving Pass@1 over default execution in 4 of 5 agent--model settings. Cost savings hold in every repository under each tested agent--model setting (\cref{fig:main-per-repo}), and \method variants account for every point on the cost--quality Pareto frontier at or above the default pass rate (\cref{fig:main-frontier}). On Web-Bench and the security benchmarks, it reduces cost by 16.0--21.2\% and 6.4--19.6\%, respectively, with Pass@2 and detection F1 remaining comparable to default execution. Our results show that our external cost-control framework can deliver consistent and transferable cost reductions across distinct agent harnesses while maintaining comparable task quality.

Our contributions can be summarized as follows:
\begin{enumerate}[noitemsep,topsep=0pt,leftmargin=8mm]
\item We formulate coding-agent cost optimization over sequences of tasks and define a standardized protocol for cumulative end-to-end accounting. Our cost profiling identifies repository-dependent effects of fixed strategies and three actionable sources of headroom.
\item We introduce \method, an external controller that learns environment-specific policies for micro-skill activation, tool pruning, and failure-loop intervention.
\item We evaluate \method across three popular coding agents and four benchmarks spanning 579 tasks, demonstrating cost reductions that hold in every setting we measure, with comparable task success.
\end{enumerate}
\section{Cost Profiling and Analysis of Fixed Cost-control Strategies}
\label{sec:pilot}

We conduct a retrospective profiling analysis on 60 SWE-Bench tasks from the verified split, with 10 tasks from each of 6 repositories, to test whether fixed strategies save cost consistently and whether completed traces reveal opportunities for adaptive control. As \cref{fig:pilot-profile-a} illustrates, every turn bills a persistent system-and-tool prefix together with the accumulated interaction history; caching discounts this replayed context but does not eliminate its cost. \cref{appendix:pilot} provides the detailed protocol.

\subsection{Repository-specific Effects of Fixed Strategies}
\label{sec:pilot-fixed}

We compare the agent's default execution with three fixed strategies, using the same configuration in every repository: (1) \textit{Rust Token Killer (RTK)} rewrites shell commands to compress their outputs;
(2) \textit{Serena} adds a semantic code-navigation interface; and (3) Aider's \textit{RepoMap} injects a structural map of repository symbols. Although their outputs may depend on the current task, none learns from completed executions or activates selectively based on prior experience.

\begin{figure}[t]
\centering
\begin{minipage}[c]{0.5\linewidth}
  \centering
  \includegraphics[width=\linewidth]{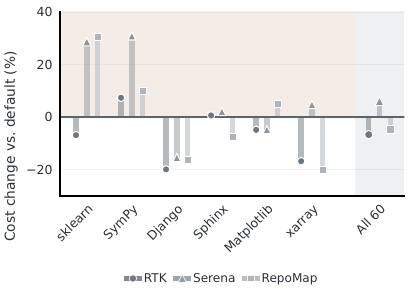}
\end{minipage}\hfill
\begin{minipage}[c]{0.46\linewidth}
  \caption{Per-repository cache-aware cost change of the three fixed strategies relative to default execution on the 60 profiled tasks (negative is cheaper). Every strategy lowers cost in some repositories and raises it in others.}
  \label{fig:pilot-strategies}
\end{minipage}
\end{figure}

\cref{fig:pilot-strategies} shows that every strategy produces mixed repository-level cost changes. For example, all three reduce cost on Django but increase it on SymPy; on scikit-learn only RTK helps, while on Sphinx only RepoMap does. These mixed effects motivate learning repository-specific policies from completed tasks.

\subsection{Three Sources of Removable Cost}

\begin{figure}[t]
\centering
\begin{subfigure}[b]{0.322\linewidth}
  \centering
  \includegraphics[width=\linewidth]{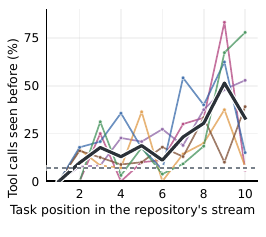}
  \caption{Repeated agent actions}
  \label{fig:pilot-headroom-a}
\end{subfigure}\hfill
\begin{subfigure}[b]{0.322\linewidth}
  \centering
  \includegraphics[width=\linewidth]{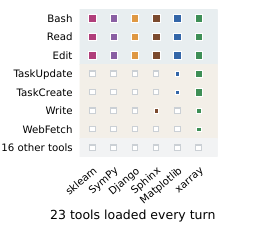}
  \caption{Loaded but irrelevant tools}
  \label{fig:pilot-headroom-b}
\end{subfigure}\hfill
\begin{subfigure}[b]{0.322\linewidth}
  \centering
  \includegraphics[width=\linewidth]{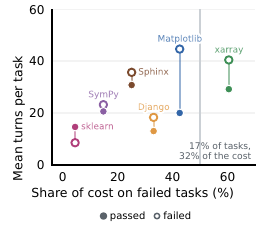}
  \caption{Cost of failed tasks}
  \label{fig:pilot-headroom-c}
\end{subfigure}
\par\vspace{3pt}
\includegraphics[width=\linewidth]{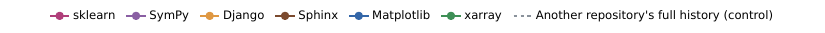}
\caption{Three trace-derived sources of cost-control headroom across the 60 profiled tasks: (a) actions repeated from earlier tasks in the same repository, with another repository's history as a control; (b) loaded tools used by each repository, with frequencies indicated by square sizes; and (c) the cost share and trajectory lengths of failed tasks.}
\label{fig:pilot-headroom}
\end{figure}

\cref{fig:pilot-headroom} identifies three potential sources of cost reduction in traces based on unmodified agent runs.

\paragraph{Operational work repeats within a repository.}
Actions that recur in at least half of a repository's tasks account for 12--28\% of the bytes its tool calls send and receive (\cref{appendix:pilot} describes how calls are matched to actions). The fraction of calls matching an action from an earlier task rises from 14\% for the second and third tasks to 30\% once at least 5 tasks precede the current one (\cref{fig:pilot-headroom-a}). Against another repository's complete history, only 7\% of calls match, primarily through generic setup and testing commands. This motivates learning repository-specific operational knowledge and injecting it only when relevant.

\paragraph{Most loaded tools remain unused.}
The agent is shown 23 tool schemas on every turn, yet no repository invokes more than 7 (\cref{fig:pilot-headroom-b}). Of these tools, 16 are never called and 3 account for 98\% of all calls; the remaining tools are sparsely and unevenly used across repositories. A fixed reduced interface could therefore remove shared overhead but also suppress a locally useful tool, motivating evidence-based pruning that can recover a pruned tool.

\paragraph{Failed trajectories consume disproportionate cost.}
The 10 failed executions constitute 17\% of the tasks but 32\% of total spend (\cref{fig:pilot-headroom-c}). A universal turn limit cannot identify them: Matplotlib failures are much longer than successful runs (45 versus 20 turns), whereas Sphinx failures run only slightly longer than its successful runs (36 versus 31). In total, 45 trajectories repeat an earlier failure signature, the normalized first error line of a failing output (\cref{appendix:mechanisms}), and 51\% of the 60 tasks' spend falls after a trajectory's first repetition. Repetition alone does not mark failure, however: 36 of the 50 successful runs also repeat a signature, and more than half of the spend after a repetition falls in runs that go on to succeed. A guard must therefore separate persistent loops from recoverable ones, for instance by comparing a run's length with the repository's successful runs, and should redirect before it terminates.

Together, these observations motivate three adaptive decisions, which \cref{sec:method} implements as meta-strategies whose policies are learned separately within each environment: which operational knowledge to inject, which persistently irrelevant tools to prune, and when to intervene in an unproductive failure loop.

\section{\method}
\label{sec:method}

\begin{figure}[t]
\centering
\includegraphics[width=\linewidth]{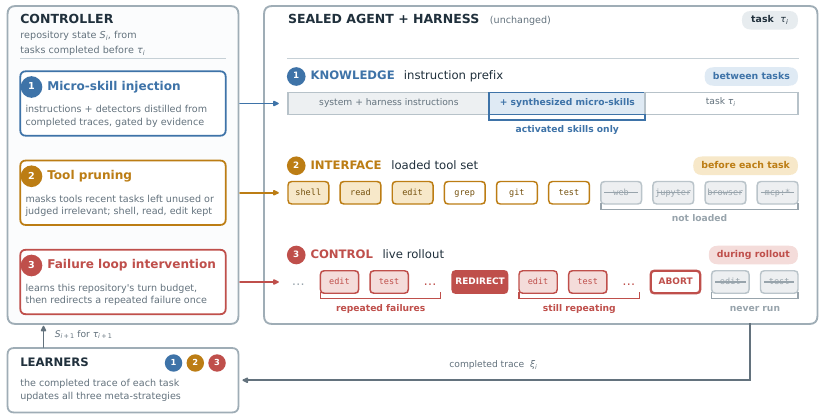}
\caption{\method applies the same three meta-strategies across repositories, while completed traces refine repository-specific policies over instructions, loaded tools, and live intervention.}
\label{fig:overview}
\end{figure}

We introduce \method, an external controller that applies three repository-agnostic meta-strategies to learn repository-specific policies from completed traces. Without modifying the underlying agent, it acts through the instruction prefix, loaded tools, and live event stream to synthesize and selectively activate operational micro-skills, prune unnecessary tools, and intervene in unproductive failure loops (\cref{fig:overview}). Throughout, a \emph{meta-strategy} is one of these three environment-agnostic procedures, a \emph{policy} is the environment-specific decision it learns, and a \emph{fixed strategy}, such as RTK or RepoMap, applies one configuration in every environment.

Our unit of optimization is a chronological task stream $\tau_1,\ldots,\tau_n$ within an environment. By default, an agent runs task $\tau_i$ under an instruction prefix $P_i$, the harness's system instructions followed by the description of $\tau_i$, with every tool in the harness's catalog loaded. When the controller is active, it maintains repository state $S_{i-1}$, which records the micro-skills and supporting evidence, tool-use history, and turn statistics accumulated from the completed tasks $\tau_1,\ldots,\tau_{i-1}$, with $S_0$ empty. Before $\tau_i$ starts, the controller $\mathcal{C}$ determines the set of active policies based on the task description and current state: 
\begin{equation}
(\Delta P_i,\mathcal{U}_i,B_i)=\mathcal{C}(\tau_i,S_{i-1}).
\label{eq:policy}
\end{equation}
The policies given by $\mathcal{C}$ are the set $\Delta P_i$ of micro-skills activated at task $i$, which depends only on $S_{i-1}$ and is rendered as a short manual inserted into $P_i$ between the harness instructions and the task description; the loaded tool set $\mathcal{U}_i$, restricted to catalog tools that remain plausibly useful; and a turn budget $B_i$ that calibrates when an in-flight failure-loop guard may redirect or terminate the rollout. The agent then runs $\tau_i$ under these policies, producing a trajectory $\xi_i$, from which the controller updates the repository state to $S_i$. The controller may also leave the agent's default configuration in place by activating no micro-skills, keeping the full tool catalog loaded, or never intervening. Its objective is to minimize mean per-task cost, including its own calls, while keeping success near default execution (\cref{appendix:formulation}, \cref{eq:objective}).

\subsection{Meta-strategy I: Experience-grounded Micro-skills}
\label{sec:injection}

\paragraph{Proposal and evidence collection.}
The controller periodically condenses a bounded window of completed trajectories into compact digests of operational evidence, such as command templates, test invocations, edited file names, and encountered errors. From these digests, an LLM \emph{proposer} drafts a few candidate micro-skills; its tokens, like those of the controller's other LLM calls, are charged to the task that triggers the call. Each candidate $f$ pairs a simple operational instruction with executable detectors, which are regular expressions or built-in predicates that test the skill against the actual task trajectory. After each task $\tau_i$, the controller runs these detectors on the shell commands and outputs in its trajectory $\xi_i$. A \emph{scope} detector checks whether $\xi_i$ \emph{exercises} the skill, that is, whether the situation it addresses arose at all; a skill on reproducing issues, for example, is exercised only if the agent attempts to write a reproduction script. An exercised task \emph{contradicts} the skill if it shows the instruction was wrong, such as an import error for a package the skill says is installed, and otherwise \emph{confirms} it. A \emph{need} detector records whether the task needed the skill: the agent either incurred the targeted waste, such as an unnecessary reinstall, or used the stated fact, such as a native test runner; without a need detector, every exercised task counts as needed. These records enter $S_i$ and decide which skills are activated or retired (examples in \cref{appendix:skill-examples}).

\paragraph{Activation and retirement.}
A candidate is activated, entering $\Delta P_i$, only while two conditions hold. It must have been confirmed on several tasks whose edited file sets do not overlap, so that its evidence reflects repository-level rather than issue-local recurrence. Its need rate, the fraction of exercising trajectories that needed it, must also stay at or above a threshold. The first condition asks whether a skill is reliable, the second whether repeatedly presenting it is worthwhile. The support of $f$ at task $i$ is
\begin{equation}
q_i(f)=\sum_{j<i}\gamma^{\,i-j}\Bigl(\1[\xi_j\ \text{confirms}\ f]-\lambda\,\1[\xi_j\ \text{contradicts}\ f]\Bigr),
\label{eq:support}
\end{equation}
where $\1[\cdot]$ is the indicator, the decay $\gamma<1$ discounts older evidence, and the weight $\lambda>1$ lets one contradiction outweigh one confirmation. A skill that has met the confirmation condition retires once $q_i(f)<0$, and its identifier is never activated again. The asymmetry is deliberate, because the two errors are not symmetric in cost: a skill that should have been retired is re-billed on every later prefix, whereas a skill retired too eagerly can be proposed again under a new identifier. A skill thus moves from candidate to confirmed to retired (\cref{tab:skill-states}); a confirmed skill below the need threshold is withheld from $\Delta P_i$ but not retired.

\subsection{Meta-strategy II: Tool Pruning}
\label{sec:untooling}

Tool pruning reduces the recurring cost of the agent's tool interface by omitting schemas unlikely to help with the current task. The controller always retains core tools for shell execution and file reading, writing, and editing. Other tools are removed after a fixed number of consecutive completed tasks have left them uncalled. In addition, an LLM judge assesses the relevance of each never-called tool not yet removed, using the tool's description and the description of $\tau_i$, in which links to the repository's pull requests, issues, commits, and files are masked. A tool is removed when it meets the unused-history criterion or the judge deems it irrelevant; the remaining tools form $\mathcal{U}_i$ and are loaded for task $\tau_i$. To recover from mistaken exclusions, the controller periodically restores the full tool catalog for a re-audit. A tool called during a re-audit is never masked again, since only never-called tools are masked.

\subsection{Meta-strategy III: Failure-loop Intervention}
\label{sec:spiral}

The controller also derives a repository-specific turn budget from prior trajectories. Let $|\xi_j|$ denote the number of turns in trajectory $\xi_j$, and let $y_j\in\{0,1\}$ indicate whether task $\tau_j$ succeeded. Let $Q_p$ denote the empirical $p$-th percentile of the turn counts of successful prior tasks. The budget for $\tau_i$ is
\begin{equation}
B_i=\beta\cdot Q_p\bigl(\{|\xi_j| : j<i,\ y_j=1\}\bigr),
\label{eq:budget}
\end{equation}
with $p$ high and a margin $\beta>1$. Here $y_j$ comes from the benchmark grader, but in realistic settings any comparable success signal can be used. The budget is a reference for identifying unusually long trajectories, not by itself a stopping point. After each tool call, a rule-based check looks for repeated failure signatures, dense error output, or a trajectory that has exceeded $B_i$, without making a model call. When triggered, an LLM judge reviews a compact digest of recent actions and running counters, including commands, edited file names, and error signatures, to distinguish an unproductive loop from a difficult task that is still progressing. A confident stuck verdict first redirects the agent with a summary of its recent actions and a request to try a different approach. If the failure pattern recurs and the judge again finds the agent stuck, the controller terminates the rollout. Productive or uncertain verdicts let it continue past $B_i$, up to the forced stop below. As a backstop that the judge cannot veto, a forced stop ends the rollout once it reaches $\kappa B_i$ turns, for a fixed $\kappa>1$, provided a strict stall rule holds: edit churn, dense errors, and a repeated failure signature at once (\cref{appendix:mechanisms}).

\cref{appendix:protocol} gives the hyperparameter values, and \cref{appendix:mechanisms} the implementation details.
\section{Experiments}
\label{sec:experiments}

We compare \method with default execution, reduced reasoning effort, and fixed strategies on cost and task success across repositories and agent--model settings, then test transfer to web development and security auditing.

\subsection{Setup}
\label{sec:setup}

\paragraph{Agents, benchmarks, and methods.}
We evaluate five agent--model settings: Claude Code with Sonnet~5 and with Opus~4.8, Copilot CLI with Sonnet~5, and Codex with GPT-5.6 Terra and GPT-5.6 Sol. Every setting runs the same 277 SWE-Bench tasks from 6 repositories~\citep{jimenez2024swebench}: scikit-learn (60), SymPy (60), Django (40), Sphinx (44), Matplotlib (34), and xarray (39). These tasks span both the verified split and harder tasks from the full split. Within a repository, tasks run in chronological order, so the controller only learns from work that came earlier in that repository's history. Three additional benchmarks assess generalizability to other domains (\cref{sec:generalization}): 102 sequential Web-Bench tasks from 5 projects~\citep{xu2025webbench}, and 100 security tasks each from \href{https://github.com/Samsung/CredData}{CredData} (credential detection) and ReposVul (vulnerability detection) \citep{wang2024reposvul}. \cref{appendix:benchmarks} describes how we selected the tasks and streams for each benchmark. In every setting, default execution is the unmodified agent at high reasoning effort, with no external control. We compare it against agent-native execution at medium and low effort, the two fixed strategies RTK and RepoMap, \method, and \method combined with RTK or RepoMap. Serena is left out because tool pruning could remove its unused server and so confound the comparison (\cref{appendix:pilot}); it was also the only fixed strategy that raised pooled cost in cost profiling (\cref{fig:pilot-strategies}).

\paragraph{Evaluation metrics.}
Our quality metric is Pass@1 under the official grader on SWE-Bench, cumulative Pass@2 on Web-Bench, as recommended by the original benchmark, and detection F1 on the security benchmarks (more details are included in \cref{appendix:benchmarks}); efficiency is mean cache-aware cost (\cref{appendix:formulation}, \cref{eq:cost}), which charges every auxiliary controller call to the task that triggered it. \cref{appendix:protocol} lists the controller configuration shared by all benchmarks.

\subsection{SWE-Bench Results}
\label{sec:swebench}

\cref{tab:main} shows that \method alone reduces mean cache-aware cost by 18.4--30.9\% across all five settings. Pass@1 improves in four settings and only falls by 1.0 percentage point on Opus 4.8. Lowered reasoning effort instead reduces Pass@1 by 1.8--10.3 percentage points, and RTK or RepoMap alone saves inconsistently. \method can also be combined with either: both combinations reduce cost in all five settings, and on Claude Code with Opus~4.8, adding RTK raises the saving from 30.9\% to 33.2\% with Pass@1 above default.

\begin{table}[t]
\centering
\caption{SWE-Bench cost and Pass@1 by method. Cost is the paired relative change from high-effort default execution and includes all controller overhead; Pass@1 is the absolute rate. Every setting runs the full six-repository set with three seeds. All values are percentages; subscripts give the standard deviation across the three seeds. Bold marks, per setting, the highest Pass@1 and the lowest cost among methods whose Pass@1 exceeds default execution; grey costs come with a Pass@1 below default execution.}
\label{tab:main}
\scriptsize
\setlength{\tabcolsep}{2pt}
\resizebox{\linewidth}{!}{\providecommand{\sd}[1]{_{\pm #1}}
\begin{tabular}{lcccccccccc}
\toprule
& \multicolumn{4}{c}{Claude Code} & \multicolumn{4}{c}{Codex} & \multicolumn{2}{c}{Copilot CLI} \\
\cmidrule(lr){2-5}\cmidrule(lr){6-9}\cmidrule(lr){10-11}
Method & \multicolumn{2}{c}{Sonnet 5} & \multicolumn{2}{c}{Opus 4.8} & \multicolumn{2}{c}{GPT-5.6 Terra} & \multicolumn{2}{c}{GPT-5.6 Sol} & \multicolumn{2}{c}{Sonnet 5} \\
\cmidrule(lr){2-3}\cmidrule(lr){4-5}\cmidrule(lr){6-7}\cmidrule(lr){8-9}\cmidrule(lr){10-11}
& $\Delta$ Cost $\downarrow$ & Pass@1 $\uparrow$ & $\Delta$ Cost $\downarrow$ & Pass@1 $\uparrow$ & $\Delta$ Cost $\downarrow$ & Pass@1 $\uparrow$ & $\Delta$ Cost $\downarrow$ & Pass@1 $\uparrow$ & $\Delta$ Cost $\downarrow$ & Pass@1 $\uparrow$ \\
\midrule
Default & --- & $74.7\sd{1.3}$ & --- & $74.0\sd{0.6}$ & --- & $73.8\sd{1.0}$ & --- & $80.1\sd{0.8}$ & --- & $75.0\sd{1.5}$ \\
Default (medium effort) & \textcolor{black!45}{$-26.6\sd{1.4}$} & $70.0\sd{1.8}$ & \textcolor{black!45}{$-16.8\sd{2.1}$} & $72.2\sd{0.4}$ & \textcolor{black!45}{$-25.7\sd{2.0}$} & $69.6\sd{1.2}$ & \textcolor{black!45}{$-28.9\sd{1.9}$} & $76.7\sd{0.7}$ & \textcolor{black!45}{$-25.4\sd{2.4}$} & $72.6\sd{1.2}$ \\
Default (low effort) & \textcolor{black!45}{$-45.5\sd{1.7}$} & $65.1\sd{0.4}$ & \textcolor{black!45}{$-31.0\sd{0.8}$} & $70.4\sd{0.4}$ & \textcolor{black!45}{$-33.0\sd{2.2}$} & $65.3\sd{1.4}$ & \textcolor{black!45}{$-43.9\sd{1.8}$} & $69.8\sd{0.4}$ & \textcolor{black!45}{$-41.9\sd{1.8}$} & $65.9\sd{0.8}$ \\
RTK & \textcolor{black!45}{$-1.1\sd{2.9}$} & $73.1\sd{0.6}$ & $-3.6\sd{3.5}$ & $74.0\sd{1.4}$ & $-4.7\sd{2.1}$ & $74.3\sd{1.1}$ & \textcolor{black!45}{$+4.5\sd{2.5}$} & $79.2\sd{1.2}$ & $-8.9\sd{3.7}$ & $76.9\sd{1.0}$ \\
RepoMap & \textcolor{black!45}{$-1.4\sd{1.9}$} & $74.1\sd{1.5}$ & $+6.5\sd{2.7}$ & $74.6\sd{1.5}$ & \textcolor{black!45}{$+3.1\sd{2.4}$} & $70.0\sd{1.8}$ & \textcolor{black!45}{$-6.1\sd{2.6}$} & $78.7\sd{1.6}$ & \textcolor{black!45}{$-3.3\sd{2.7}$} & $74.9\sd{1.3}$ \\
\midrule
\method & $\mathbf{-23.6}\sd{1.8}$ & $76.0\sd{1.2}$ & \textcolor{black!45}{$-30.9\sd{2.1}$} & $73.0\sd{0.6}$ & $\mathbf{-24.7}\sd{1.2}$ & $74.7\sd{1.0}$ & $\mathbf{-26.1}\sd{2.1}$ & $\mathbf{81.7}\sd{0.7}$ & $\mathbf{-18.4}\sd{1.5}$ & $\mathbf{77.6}\sd{1.3}$ \\
\method+RTK & $-23.5\sd{1.2}$ & $76.4\sd{2.4}$ & $\mathbf{-33.2}\sd{1.7}$ & $74.8\sd{1.2}$ & $-23.8\sd{1.5}$ & $\mathbf{75.2}\sd{0.9}$ & $-23.3\sd{1.9}$ & $81.3\sd{1.4}$ & $-13.2\sd{2.2}$ & $77.4\sd{1.1}$ \\
\method+RepoMap & $-18.3\sd{3.2}$ & $\mathbf{76.7}\sd{1.1}$ & $-24.9\sd{3.9}$ & $\mathbf{75.1}\sd{0.7}$ & \textcolor{black!45}{$-21.9\sd{3.2}$} & $73.5\sd{1.6}$ & \textcolor{black!45}{$-29.2\sd{2.2}$} & $79.0\sd{1.3}$ & $-6.3\sd{1.6}$ & $76.8\sd{0.4}$ \\
\bottomrule
\end{tabular}
}
\end{table}

\begin{figure}[t]
\centering
\begin{subfigure}[b]{0.485\linewidth}
  \centering
  \includegraphics[width=\linewidth]{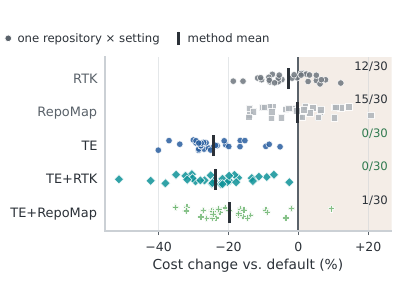}
  \caption{Per-environment consistency}
  \label{fig:main-per-repo}
\end{subfigure}\hfill
\begin{subfigure}[b]{0.485\linewidth}
  \centering
  \includegraphics[width=\linewidth]{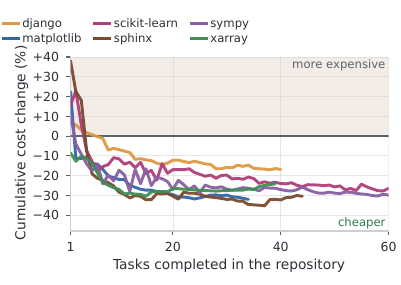}
  \caption{Cumulative cost change}
  \label{fig:main-amortization}
\end{subfigure}
\caption{SWE-Bench results relative to default execution. (a) Cost change in each of the 30 repository--setting cells. (b) Cumulative cost change of \method over each repository's task stream, averaged over the two Claude Code settings. The cost--quality frontier is in \cref{fig:main-frontier}.}
\label{fig:main-results}
\end{figure}

\paragraph{The savings are consistent.}
Pooled savings can conceal cost increases in individual environments, as the fixed strategies in \cref{sec:pilot} illustrate. \cref{fig:main-per-repo} reports cost changes for each of the six repositories under each of the five agent--model settings, yielding 30 repository--setting cells. \method reduces cost in all 30, by 5.2--40.1\%, as does \method{}+RTK. \method{}+RepoMap reduces cost in 29 of 30. By comparison, RTK alone increases cost in 12 cells and RepoMap alone in 15.

\paragraph{Adaptation expands the Pareto frontier.}
\cref{fig:main-frontier} places each method’s cost and Pass@1 change relative to default execution in its agent--model setting. Reduced reasoning effort achieves large savings with substantial losses in Pass@1, while RTK and RepoMap stay closer to default and move in mixed directions. \method variants account for every point on the Pareto frontier that maintains or improves the default pass rate, showing that adaptive control reaches a more favorable cost--quality tradeoff across settings.

\paragraph{Savings accumulate over the task stream.}
\method begins each repository without prior traces, so its benefit may take several tasks to emerge. \cref{fig:main-amortization} plots the cumulative cost change of \method (without integrations) relative to default execution within each repository, averaged over the two Claude Code settings. All repositories except xarray initially cost more under \method, but all become cheaper by the sixth task and remain cheaper thereafter, ending with cumulative savings of 16.9--32.0\%. 

\paragraph{Ablations.}
We run each meta-strategy alone. In the two Claude Code settings, tool pruning carries the largest single share, $-13.2\%$ and $-17.7\%$, because it acts on every turn of every task, with micro-skill injection close behind at $-11.8\%$ and $-14.3\%$ and the failure-loop guard in single digits. Pass@1 stays within 3 points of default for every variant, so none trades task success for cost, and none alone reproduces the full stack's saving (\cref{tab:ablation-main}; activity counts in \cref{appendix:ablation}).

\subsection{Generalization to Diverse Task Structures}
\label{sec:generalization}

To assess generalization beyond SWE-Bench, we evaluate \method on sequential web development (Web-Bench) and two security-auditing benchmarks (CredData and ReposVul). Web-Bench typically involves smaller codebases and workspaces than SWE-Bench, but its tasks remain within one continuous project and naturally inherit earlier workspace changes. In the security benchmarks, each task targets a different repository under a fixed audit prompt, so the procedure recurs without repository continuity. Because no repository recurs, we disable repository-specific micro-skill injection, leaving only two meta-strategies active.

\cref{tab:transfer} shows cost reductions of 16.0--21.2\% on Web-Bench and 16.5--19.6\% on CredData. The latter result shows that the remaining meta-strategies can still achieve substantial savings across new repositories under a shared audit procedure. Reductions are much smaller on ReposVul, at 6.4--9.3\%, where shorter trajectories leave less removable cost. The savings thus generalize across these workloads, while their magnitude varies with the reusable structure and cost-reduction headroom within the trajectories.

\begin{table}[t]
\centering
\caption{Cost and quality relative to default execution (\%); subscripts are standard deviations over three seeds. (a) Each meta-strategy alone and the full stack on SWE-Bench with Claude Code. (b) \method (TE) on Web-Bench (102 tasks; tasks after an early stop fail; cost paired over tasks both arms reached) and the security benchmarks (100 tasks each); Claude Code averages Sonnet~5 and Opus~4.8.}
\label{tab:ablation-transfer}
\scriptsize
\setlength{\tabcolsep}{3.6pt}
\begin{subtable}[t]{0.47\linewidth}
  \centering
  \caption{Meta-strategy ablation}
  \label{tab:ablation-main}
  \providecommand{\sd}[1]{_{\pm #1}}
\begin{tabular}{llrr}
\toprule
 & & \multicolumn{2}{c}{SWE-Bench} \\
\cmidrule(lr){3-4}
Model & Method & $\Delta$ Cost $\downarrow$ & Pass@1 $\uparrow$ \\
\midrule
Sonnet 5 & Default & --- & $74.7\sd{1.3}$ \\
 & Micro-skills & $-11.8\sd{2.1}$ & $77.3\sd{0.4}$ \\
 & Tool pruning & $-13.2\sd{1.3}$ & $74.3\sd{0.5}$ \\
 & Failure-loop guard & $-7.1\sd{2.2}$ & $76.2\sd{0.9}$ \\
 & Combined & $-23.6\sd{1.8}$ & $76.0\sd{1.2}$ \\
\addlinespace
Opus 4.8 & Default & --- & $74.0\sd{0.6}$ \\
 & Micro-skills & $-14.3\sd{2.3}$ & $74.2\sd{1.1}$ \\
 & Tool pruning & $-17.7\sd{1.5}$ & $73.5\sd{0.6}$ \\
 & Failure-loop guard & $-5.4\sd{1.8}$ & $72.6\sd{0.5}$ \\
 & Combined & $-30.9\sd{2.1}$ & $73.0\sd{0.6}$ \\
\bottomrule
\end{tabular}

\end{subtable}\hfill
\begin{subtable}[t]{0.51\linewidth}
  \centering
  \caption{Generalizability to other benchmarks}
  \label{tab:transfer}
  \providecommand{\sd}[1]{_{\pm #1}}
\begin{tabular}{llrrr}
\toprule
 & & & \multicolumn{2}{c}{Quality $\uparrow$} \\
\cmidrule(lr){4-5}
Workload & Agent & $\Delta$ Cost $\downarrow$ & Default & TE \\
\midrule
Web-Bench & Claude Code & $-21.2\sd{1.3}$ & $69.3\sd{1.4}$ & $73.3\sd{1.6}$ \\
(Pass@2) & Codex & $-16.0\sd{1.5}$ & $64.0\sd{1.5}$ & $66.5\sd{0.7}$ \\
 & Copilot CLI & $-18.7\sd{1.9}$ & $60.8\sd{2.2}$ & $59.3\sd{1.9}$ \\
\addlinespace[6.5pt]
CredData & Claude Code & $-19.6\sd{0.7}$ & $67.9\sd{0.5}$ & $70.1\sd{0.8}$ \\
(F1) & Codex & $-16.5\sd{1.0}$ & $68.0\sd{0.4}$ & $69.2\sd{0.3}$ \\
 & Copilot CLI & $-18.3\sd{1.4}$ & $68.2\sd{0.4}$ & $67.7\sd{0.7}$ \\
\addlinespace[6.5pt]
ReposVul & Claude Code & $-7.5\sd{1.6}$ & $17.5\sd{1.1}$ & $19.6\sd{0.8}$ \\
(F1) & Codex & $-6.4\sd{1.2}$ & $19.4\sd{0.9}$ & $20.1\sd{0.6}$ \\
 & Copilot CLI & $-9.3\sd{1.6}$ & $21.6\sd{1.4}$ & $22.5\sd{0.9}$ \\
\bottomrule
\end{tabular}

\end{subtable}
\end{table}

\section{Related Work}
\label{sec:related}

\paragraph{Cost reduction for LLMs and coding agents.}
A classic line of work lowers inference cost by choosing which model answers each query: cascades escalate from cheaper to stronger models until an answer is judged reliable~\citep{chen2023frugalgpt,yue2024cascades,aggarwal2024automix}, and routers predict the cheapest adequate model before generation~\citep{ding2024hybrid,ong2024routellm}. These methods trade answer quality for cost on each query, whereas \method keeps the model and effort fixed and removes cost that recurs within an environment. For coding agents, existing methods reduce cost within a trajectory by controlling context, interfaces, or rollout length. Context-control methods compress prompts~\citep{jiang2023llmlingua}, shell output with RTK, tool observations~\citep{kovacs2026squeez}, or accumulated trajectories~\citep{xiao2025agentdiet}; others narrow retrieval through repository maps, knowledge graphs~\citep{vogel2026codebasememory}, structure-aware scouting~\citep{li2026fastcode}, or trained retrieval agents~\citep{bashir2026context1}. Interface-level methods restrict or retrieve candidate tools~\citep{paramanayakam2024lessismore,du2024anytool}, enrich tool interfaces~\citep{liu2025align}, or encourage minimal implementations through plugins such as \href{https://github.com/dietrichgebert/ponytail}{Ponytail}. Runtime methods provide corrective feedback~\citep{zhu2025agentdebug}, early-exit checks~\citep{lu2025runaway}, budget estimates~\citep{lin2026bagen}, or hidden-state failure detection~\citep{ruan2026doomed}. Methods deployable through an agent’s exposed interfaces are typically fixed and task-local; our cost profiling (\cref{sec:pilot}) finds that their effects vary across repositories. \method instead uses completed traces to adapt the instruction prefix, tool set, and stopping policy to each environment.

\paragraph{Cross-task experience for coding agents.}
Recent benchmarks have started to evaluate agents across continual repair, related-task context reuse, and streaming task sequences~\citep{joshi2025swebench-cl,zhu2026swecontext,yan2026agentstream}. Prior methods reuse completed executions as verbal lessons or workflows~\citep{shinn2023reflexion,zhao2024expel,wang2024workflowmemory}, or derive repository knowledge from commit histories and repair trajectories~\citep{wang2026repomemory,chen2025sweexp}. A complementary line of work packages experience into reusable capabilities, including executable skill and tool libraries~\citep{wang2023voyager,cai2024large,yuan2024craft}, verified programmatic skills~\citep{wang2025inducing}, refined skill APIs~\citep{zheng2025skillweaver}, and evolving test-time libraries~\citep{xu2026test}. Recent systems further automate skill construction and evolution~\citep{wang2026skillx,ni2026trace2skill,ma2026skillgen,zhang2026coevoskills,chen2026skillcraft,li2026codeskill}. Because indiscriminate repository context can increase cost without generally improving success~\citep{gloaguen2026agentsmd}, such experience must be reused selectively. Building on this principle, \method uses cross-task experience to learn cost-control policies applied through agents’ exposed interfaces.
\section{Limitations and Conclusion}
\label{sec:discussion}

\paragraph{Limitations.}
Each meta-strategy can still make an incorrect decision: a micro-skill may become stale, a pruned tool may become necessary, and a difficult but productive task may be mistaken for a failure loop. Each decision, however, comes with a recovery path that limits the damage: a stale skill is retired once a trace contradicts it, a wrongly pruned tool returns at the next re-audit, and the failure-loop intervention redirects before it stops and spares uncertain cases. Our experiments cover three widely used coding agents but not more minimal harnesses such as \href{https://github.com/earendil-works/pi}{Pi}; since \method relies only on the prompt, tool configuration, and event stream, supporting such a harness requires an adapter rather than new meta-strategies. Finally, our SWE-Bench streams consist of related but independent fixes rather than actual user sessions. Continuous sessions should offer more reusable experience (\cref{appendix:discussion}), so interactive coding workflows are a natural next test.

\paragraph{Conclusion.}
Coding-agent costs accumulate across tasks in recurring environments, which isolated-task evaluations overlook. \method shows that an external controller can learn from completed work to make an unchanged agent cheaper while largely preserving task success, across software engineering, web development, and security auditing.

{
    \small
    \bibliography{main}
    \bibliographystyle{iclr2027_conference}
}

\appendix
\crefalias{section}{appendix}
\crefalias{subsection}{appendix}
\crefalias{subsubsection}{appendix}
\section{Extended Discussion}
\label{appendix:discussion}

\paragraph{\method as one instance of a broader class.}
\method occupies one point in a larger design space: external controllers that apply fixed, environment-agnostic meta-strategies and let completed traces set environment-specific policies. Each of our meta-strategies follows the same pattern, pairing a cheap signal from traces with a conservative action and a way to recover from mistakes: micro-skills are confirmed by detectors and retired on contradiction, pruned tools return through re-audits, and the failure-loop guard redirects before it stops. The same pattern applies to other controls that harnesses expose, such as choosing the reasoning effort or model for each task, compacting context at learned points, or reusing environment setup across tasks, and each strategy could be specialized further to a deployment, for example with longer histories, micro-skills shared across related repositories, or success signals from users rather than benchmark graders. Because the controller needs no model access, retraining, or change to the agent, it can follow agents as they evolve. We therefore see our results as evidence that this direction is promising and underexplored, not as its ceiling.

\paragraph{Generalization without per-environment tuning.}
A single configuration (\cref{tab:config}) serves three harnesses, five agent--model settings, and four benchmarks; the only benchmark-specific changes follow from the structure of the streams (\cref{appendix:protocol}). That the savings hold in all 30 SWE-Bench repository--setting cells without tuning suggests that the meta-strategies capture regularities of agentic work, namely recurring operational facts, unused interfaces, and unproductive loops, rather than artifacts of one agent or dataset. The shared configuration is a default rather than a constraint, and its parameters also serve as knobs with a predictable direction: fewer required confirmations or a lower need gate activate micro-skills sooner, a shorter unused streak or a longer re-audit period keeps more tools pruned, and a smaller budget margin $\beta$, a lower judge confidence threshold, or a smaller $\kappa$ lets the guard intervene earlier. Each adjustment trades additional savings against a higher risk of a wrong decision, so users can move along the cost--quality frontier to suit their priorities, much as with reasoning effort, whose savings come with larger losses in Pass@1 (\cref{fig:main-frontier}). SWE-Bench is also a conservative test of experience reuse. Consecutive tasks are independent: each starts from a fresh checkout of its own base commit, none inherits the agent's earlier edits, and each stream spans 1.9 to 8.5 years of its repository's history, over which layouts, test commands, and dependencies drift. A developer's session instead returns to a codebase that changes incrementally, often through the agent's own work, so more of what one task teaches should remain valid for the next. Web-Bench, whose tasks inherit the workspace, is closer to that setting, and \method saves 16.0--21.2\% there even though its streams have only 20 or 21 tasks.

\paragraph{Complementarity with fixed strategies.}
Alone, RTK and RepoMap change Pass@1 inconsistently, by $-3.8$ to $+1.9$ points across settings (\cref{tab:main}). Their combinations with \method are more informative. On Claude Code with Opus~4.8, where \method alone is 1.0 point below default, adding RepoMap or RTK raises Pass@1 to 75.1 and 74.8, above default, and a combination attains the highest Pass@1 in three of the five settings. A plausible explanation is that the strategies supply different information. Micro-skills state how to operate in a repository, such as how to run its tests, but by design say nothing about where a fix belongs; RepoMap's ranked signatures address that localization question, and RTK's condensed logs keep failure messages salient in a shorter context. The savings, however, do not simply add up: RepoMap bills its map on every turn, so \method{}+RepoMap saves less than \method alone in four of five settings, and RTK, whose own savings are inconsistent, adds to \method{}'s saving on Opus~4.8 but lowers it by up to 5.2 points elsewhere. Combining fixed and adaptive strategies thus trades some savings for quality, and which combination is preferable depends on the setting.

\paragraph{Why micro-skills.}
We call the injected knowledge micro-skills to distinguish it from two forms of reusable experience in prior work (\cref{sec:related}). Skill and tool libraries package executable code, programs, or APIs for recurring subtasks, while lessons, workflows, and repository memories distill past trajectories into natural-language guidance on how to solve tasks. A micro-skill is narrower than either: a single operational fact about one environment, such as which test runner works, where sources live, or which setup step is unnecessary. Three properties make this granularity suitable for cost control. First, it targets the cost we observe: frontier agents already solve most tasks, and much of their removable spend goes to rediscovering the same environment facts from task to task (\cref{sec:pilot}) rather than to missing problem-solving ability. Second, it is continually verifiable: detectors confirm or contradict a one-line claim about commands and outputs on every later trace without a model call, so its validity is rechecked throughout the stream rather than once when it is created, whereas whether a general lesson or procedure helped is hard to read from a trace. Third, it is cheap and low-risk: the manual is billed on every turn, so each entry must be short and earn its place, and because a micro-skill states facts rather than solutions, it carries no task-specific content from earlier tasks into later ones, and a wrong one is retired after little evidence, as Example~(c) in \cref{appendix:skill-examples} shows.

\section{Cost Profiling Protocol and Detailed Results}
\label{appendix:pilot}

Cost profiling runs Claude Code with Opus~4.8 at high reasoning effort. From each of the 6 repositories we evaluate 10 tasks, and run default execution along with each of the three fixed strategies once on the same 60 tasks. Each strategy's cost change is relative to the mean cost of default execution. \cref{tab:pilot} reports numerical per-repository results.

\begin{table}[ht]
\centering
\caption{Cost-profiling results by repository: tasks passed out of 10, and each strategy's cost change relative to default execution on the same tasks.}
\label{tab:pilot}
\small
\setlength{\tabcolsep}{4pt}
\begin{tabular}{lcrrrrrr}
\toprule
 & Default & \multicolumn{2}{c}{RTK} & \multicolumn{2}{c}{Serena} & \multicolumn{2}{c}{RepoMap} \\
\cmidrule(lr){3-4}\cmidrule(lr){5-6}\cmidrule(lr){7-8}
Repository & Pass@1 $\uparrow$ & $\Delta$ cost $\downarrow$ & Pass@1 $\uparrow$ & $\Delta$ cost $\downarrow$ & Pass@1 $\uparrow$ & $\Delta$ cost $\downarrow$ & Pass@1 $\uparrow$ \\
\midrule
sklearn & 10/10 & $-6.9\%$ & 10/10 & $+28.6\%$ & 10/10 & $+30.5\%$ & 10/10 \\
SymPy & 10/10 & $+7.3\%$ & 10/10 & $+30.8\%$ & 10/10 & $+9.8\%$ & 8/10 \\
Django & 9/10 & $-19.9\%$ & 9/10 & $-15.4\%$ & 8/10 & $-16.2\%$ & 9/10 \\
Sphinx & 6/10 & $+0.6\%$ & 8/10 & $+2.1\%$ & 6/10 & $-7.5\%$ & 8/10 \\
Matplotlib & 8/10 & $-4.9\%$ & 9/10 & $-4.8\%$ & 8/10 & $+5.0\%$ & 9/10 \\
xarray & 7/10 & $-16.8\%$ & 7/10 & $+4.7\%$ & 6/10 & $-20.2\%$ & 7/10 \\
\bottomrule
\end{tabular}

\end{table}

\paragraph{External cost-saving strategies.}
\begin{itemize}[noitemsep,topsep=2pt,leftmargin=5mm]
\item \emph{RTK} (v0.45.0) is registered as a pre-execution hook on the agent's shell tool. It rewrites each shell command to run through RTK, which condenses the command's output, such as test logs, listings, and diffs, before the agent sees it. The agent's tools and prompt are unchanged.
\item \emph{Serena} is an MCP server backed by language servers. It adds symbol-level navigation tools, such as finding a symbol's definition, listing a file's symbols, and finding references. The agent is instructed to prefer these tools over its native file reading and search, which remain available.
\item \emph{RepoMap} refers to Aider's repository map. It parses the repository's Python source, ranks files and symbols by their centrality in the cross-reference graph, biased toward identifiers mentioned in the task description, and renders the top-ranked signatures within a 2,048-token budget. The map is generated once per task and supplied as project instructions, so it becomes part of the prefix billed on every turn.
\end{itemize}
The main experiments use the same RTK version and map budget, and also test \method{}+RTK and \method{}+RepoMap to determine whether adaptive savings complement output compression and static repository context. There is no \method{}+Serena variant, because tool pruning may remove an unused Serena server and so directly confound the two methods; for the same reason, Serena is absent from \cref{tab:main}.

\paragraph{Headroom measurements.}
All three measurements read the default-execution traces and estimate exposure to removable cost, which should be interpreted as an upper bound rather than achievable savings.

\begin{itemize}[noitemsep,topsep=2pt,leftmargin=5mm]
\item \emph{Repeated actions} (\cref{fig:pilot-headroom-a}). We reduce each tool call to an action key. For shell commands, the key is the command lower-cased, with whitespace collapsed, any leading directory change removed, and only the first 6 tokens kept. For file tools, the key is the tool name and the file's base name. Within each repository, tasks are taken in chronological order, and for each task we measure the share of its tool calls whose key already appeared in an earlier task of the same repository. The cross-task control measures the same share against the complete history of each other repository, averaged over the five. 
\item \emph{Unused tools} (\cref{fig:pilot-headroom-b}). We compare the tool catalog the agent receives at session start, which has 23 tools and is the same in every repository, with the tool calls it actually makes, counted per repository.
\item \emph{Failed tasks} (\cref{fig:pilot-headroom-c}). We measure the share of default spend that goes to tasks that fail, and the mean turn count of passed and failed tasks. For the spend after a trajectory first repeats a failure signature, each run's cost is spread evenly over its tool calls, which understates late spend, since every turn re-bills the context accumulated before it. 
\end{itemize}

\section{Controller Configuration}
\label{appendix:protocol}

\cref{tab:config} lists the controller configuration, and \cref{appendix:codex} describes how each harness implements it. Within a benchmark, every repository and project uses the same values, and none is tuned per repository, project, or stream. All values were fixed before the main runs, in earlier development runs on the same six SWE-Bench repositories. These values serve six SWE-Bench repositories whose streams range from 34 to 60 tasks and whose default pass rates, turn counts, and tool usage differ widely, as well as five Web-Bench projects that span Web standards and frameworks.

The settings differ across benchmarks only where the structure of the stream requires it. Web-Bench streams have 20 or 21 tasks, too few for the percentile of passed turn counts to warm up before most of the stream has run. The failure-loop guard therefore starts from a pre-registered budget of 30 turns until 5 tasks have passed. On the security benchmarks every task is a different repository, so a repository-specific micro-skill never has a later task in the same repository to apply to; micro-skill injection is disabled. Their tasks are also read-only audits under one fixed brief, so the unused-tool history reflects the audit procedure rather than any single repository, and the judge assesses each candidate against that brief.

\begin{table}[ht]
\centering
\caption{Controller configuration, shared by all three harnesses.}
\label{tab:config}
\small
\setlength{\tabcolsep}{4pt}
\begin{tabularx}{0.8\linewidth}{@{}>{\raggedright\arraybackslash}p{3.2cm}>{\raggedright\arraybackslash}X@{}}
\toprule
Config & Value \\
\midrule
\multicolumn{2}{@{}l}{\emph{Micro-skill injection}} \\
Skill sources & proposer \\
Proposer cadence & every 5 tasks \\
Confirmation & 3 confirmations on non-overlapping edited files \\
Need gate & rate $\geq 0.4$ over $\geq 4$ exercises \\
Support $\gamma$, $\lambda$ & $0.8$, $5$ \\
\addlinespace
\multicolumn{2}{@{}l}{\emph{Tool pruning}} \\
Pruning unit & tool (Codex: capability package) \\
Removal & unused for 3 tasks, plus per-task judge \\
Recovery & full catalog every 5 tasks \\
\addlinespace
\multicolumn{2}{@{}l}{\emph{Failure-loop intervention}} \\
Budget $B_i$ & $1.25\times$ 90th percentile of passed turns \\
Stuck detection & LLM judge, confidence $0.7$ \\
Redirects & 1, then stop \\
Forced stop $\kappa$ & $2.5$, if the strict stall rule holds \\
\bottomrule
\end{tabularx}
\end{table}

\section{Task and Stream Selection}
\label{appendix:benchmarks}

All methods and seeds run the same tasks in the same order, so seeds differ only in the agent's own nondeterminism.

\paragraph{SWE-Bench.}
We use the six repositories profiled in \cref{sec:pilot}, all among the largest in SWE-Bench Verified. Of the 277 tasks, 199 come from SWE-Bench Verified and 78 from the full SWE-Bench test split:
\begin{itemize}[noitemsep,topsep=2pt,leftmargin=5mm]
\item \emph{Django} (40), \emph{Matplotlib} (34), and \emph{Sphinx} (44) use only Verified tasks. Matplotlib and Sphinx use every Verified task; Django uses 40 of its 231 Verified tasks, all from the start of its Verified history (November 2016 to September 2019), so that consecutive tasks are typically days apart (median 5 days between issue dates) rather than spread over the full 6.7 years.
\item \emph{scikit-learn} (60) combines 31 Verified tasks with 29 from the full split. 
\item \emph{SymPy} (60) combines 41 Verified tasks with 19 from the full split.
\item \emph{xarray} (39) has only 22 Verified tasks, and the 13 oldest are spread thinly over 2019 to 2021. Its stream instead uses the 9 Verified tasks from April 2022 onward, interleaved in base-commit order with 30 full-split tasks, one from February 2021 and the rest from March 2022 to January 2023, so that consecutive tasks are typically days apart (median 4 days between issue dates).
\end{itemize}
We draw the extra scikit-learn and SymPy tasks from the full split to raise the difficulty of the two streams. With Verified tasks alone, nearly every method passes nearly every task in these repositories, so default execution cannot be told apart from its reduced-effort variants. The added tasks have gold patches of 61--200 changed lines and a package version that already appears in the stream; we sort candidates by patch size and sample at even spacing. Because the full split was not validated by human annotators, we checked each added task by applying its gold patch in the official image and running the official evaluation, and every one passed. Each repository's stream runs in base-commit order. Because fix commits are public, every method runs behind an answer-key guard that blocks direct access to the gold patch on GitHub (\cref{appendix:guard}).

\paragraph{Web-Bench.}
Web-Bench consists of 50 projects, each a sequence of 20 dependent tasks, and some projects open with an extra \texttt{init} task that sets up the scaffold. We run five projects as streams in the benchmark's own task order: calculator (21 tasks), table (21), expressjs (20), react (20), and zustand (20), for 102 tasks in total. Each setting runs every stream with three seeds. We chose the five for diversity across the benchmark's two halves, Web standards and Web frameworks. Calculator and table are built on plain HTML, CSS, and JavaScript, the latter around the HTML \texttt{<table>} element. React is a front-end UI framework, zustand a state-management library used with React, and expressjs a Node.js back end with a database. We also excluded projects on which default execution fails early: the official protocol stops a stream at the first task that fails both attempts, so such a project would leave too short a stream to compare methods on.

We follow the official Web-Bench evaluation protocol~\citep{xu2025webbench}, which allows each task two attempts. If the first attempt fails the project's tests, the agent retries the same task with the test errors as context. The retry matters because each task builds on the previous one: without it, a single failed attempt would end the stream and leave every later task unevaluated. We therefore report Pass@2, the fraction of tasks passed on either attempt, and our cost figures include both attempts. As on SWE-Bench, every method runs behind the answer-key guard (\cref{appendix:guard}), because the grading specifications are public.

\paragraph{CredData.}
CredData contains 333 repositories. We exclude the 19 that contain no true secret line: for those, predicting nothing scores an F1 of 1, so they measure only precision. We split the remaining 314 into quartiles by the number of true secret lines per repository, which is heavily skewed (cut points 4, 8.5, and 24.75), allocate the 100 slots in proportion (28, 22, 25, and 25), and select that many repositories from each quartile. The selected repositories contain between 1 and 799 true secret lines (median 8.5).

\paragraph{ReposVul.}
From the 1,310 instances of the Python split, we drop the 298 whose fix removes no line and so leaves nothing to detect, and the 109 that ReposVul flags as outdated. From the remaining 903 we sample 100 as for CredData, by quartile of vulnerable-line count (cut points 2, 5, and 17; 31, 21, 24, and 24 slots) with a fixed seed. The selected instances have 1 to 411 vulnerable lines (median 5). As on CredData, every task is a different repository under the same fixed audit prompt, and each seed runs all 100. We grade strictly at the line level: a reported location counts only if its file and line match a line the fixing commit removed, and precision is taken over reported lines in the vulnerable files. Prior work instead reports binary F1 over whole functions or code samples on balanced sets~\citep{lekssays2025llmxcpg}; under our metric, naming the right function but a different line scores zero. We keep the strict metric because it mirrors CredData's line-level F1 and needs no further labels.

\section{Task Stream Formulation and Cost Accounting}
\label{appendix:formulation}

\paragraph{Task-stream formalization.}
\cref{sec:method} defines the task stream, the prefix $P_i$, and the controller map $\mathcal{C}$ of \cref{eq:policy}; we reuse that notation here, writing $S_i=F(S_{i-1},\tau_i,\xi_i,y_i)$ for the state update. The trajectory alternates observations and agent actions, $\xi_i=(o_0,a_0,o_1,\ldots,a_T,o_{T+1})$, so its turn count in \cref{eq:budget} is $|\xi_i|=T+1$, and its success label is $y_i\in\{0,1\}$. Because $S_{i-1}$ is formed only from tasks preceding $i$ and the failure-loop guard observes only the prefix $\xi_{i,\leq t}$, no decision for $\tau_i$ depends on a later task in the stream.

\paragraph{Cost attribution and objective.}
For task $\tau_i$, let $N^{\mathrm{u}}_i$, $N^{\mathrm{c}}_i$, and $N^{\mathrm{o}}_i$ denote the uncached-input, cache-read, and output tokens consumed by the task agent. Let $\mathcal{H}_i$ contain every auxiliary model call attributable to the task, including micro-skill proposal, tool judging, trace judging, and redirect writing. For each auxiliary call $h$, we record the same three token classes. Controller overhead and total cache-aware cost are then
\begin{equation}
\begin{aligned}
H_i
&=
\sum_{h\in\mathcal{H}_i}
\left(
w_u N^{\mathrm{u}}_{i,h}
+w_c N^{\mathrm{c}}_{i,h}
+w_o N^{\mathrm{o}}_{i,h}
\right), \\
\cost(\tau_i;\mathcal{C})
&=
w_u N^{\mathrm{u}}_i+w_c N^{\mathrm{c}}_i+w_o N^{\mathrm{o}}_i+H_i,
\end{aligned}
\label{eq:cost}
\end{equation}
where the weights are the applicable prices, or their ratios when cost is expressed in fresh-input-token equivalents. We use provider-reported token classes, so cached context is priced as cache-read usage rather than reconstructed as fresh input.

Each auxiliary call is charged exactly once to the task boundary that invokes it. For example, a periodic post-task update is assigned to the task whose completion triggers the update. For the stream of environment $r$, the resulting cost and success are
\begin{equation}
\begin{aligned}
\overline{C}_r(\mathcal{C})
&=
\frac{1}{n}
\sum_{i=1}^{n}
\cost(\tau_i;\mathcal{C}), \\
\overline{Y}_r(\mathcal{C})
&=
\frac{1}{n}
\sum_{i=1}^{n} y_i.
\end{aligned}
\label{eq:stream-metrics}
\end{equation}
Conceptually, \method seeks to reduce amortized cost while preserving the success of default execution:
\begin{equation}
\min_\mathcal{C}\;
\mathbb{E}\!\left[\overline{C}_r(\mathcal{C})\right]
\quad\text{subject to}\quad
\mathbb{E}\!\left[\overline{Y}_r(\mathcal{C})\right]
\geq
\mathbb{E}\!\left[\overline{Y}_r(\mathrm{default})\right]
-\epsilon.
\label{eq:objective}
\end{equation}
The expectation is over stochastic agent execution. \cref{eq:objective} states the design objective rather than an evaluation-time tuning rule: we do not select $\epsilon$ after observing the results. Instead, experiments report the observed cost change and success difference relative to matched default executions.

At finalization, task-agent usage and all attributed controller calls are written to the same execution record before \cref{eq:cost} is evaluated. This shared accounting path prevents an optimized method from omitting work moved into an auxiliary call or obtaining an unreported discount through different model routing.

\section{Additional Meta-strategy Details}
\label{appendix:mechanisms}

\paragraph{Micro-skill injection.}
The proposer is called after task 4 and every 5 tasks thereafter. Each call sees digests of the repository's 8 most recent tasks, which contain no solution text or gold patches, and returns at most 4 candidates. A candidate is discarded unless it cites at least 2 tasks in the window on which its confirming detector matches. An admitted candidate's detectors are run retroactively over the window's tasks, so it enters with their evidence and can be activated at once. On an exercised task, contradiction is checked before confirmation, and a skill without a need detector counts every exercised task as needed. A skill the proposer marks as high-stakes, such as a test command or a build step, retires at its first contradiction once it has met the confirmation condition. \cref{tab:skill-states} summarizes the lifecycle. The three criteria act at different points: the confirmation count moves a candidate to confirmed, the need rate decides on each task whether a confirmed skill is injected, and the support $q_i(f)$ decides retirement. Support is checked only once a skill is confirmed, so a candidate with negative support, which is never injected, stays a candidate; if it is later confirmed, it retires after the next task unless its support has turned positive.

\begin{table}[ht]
\centering
\caption{Micro-skill lifecycle. Only a confirmed skill can enter $\Delta P_i$, and retirement is permanent.}
\label{tab:skill-states}
\small
\setlength{\tabcolsep}{4pt}
\begin{tabularx}{\linewidth}{@{}>{\raggedright\arraybackslash}p{1.6cm}>{\raggedright\arraybackslash}X>{\raggedright\arraybackslash}X@{}}
\toprule
State & Injected into $\Delta P_i$ & Leaves the state when \\
\midrule
Candidate & never & confirmed on 3 tasks with non-overlapping edited files $\rightarrow$ confirmed \\
Confirmed & on each task, if its need rate is $\geq 0.4$ over $\geq 4$ exercised tasks; otherwise withheld, and rechecked before the next task & $q_i(f)<0$, or any contradiction of a high-stakes skill $\rightarrow$ retired \\
Retired & never & never; the proposer may propose a replacement under a new identifier \\
\bottomrule
\end{tabularx}
\end{table}

\paragraph{Tool pruning.}
A tool's unused streak resets whenever it is called. Before each task, unprotected, never-called tools whose streak has reached three are masked; the other unprotected, never-called tools are sent to the judge, and those it marks irrelevant are masked too; if the judge call fails, they stay loaded. The mask is fixed for the whole task. A re-audit loads the full catalog without calling the judge.

\paragraph{Failure-loop intervention.}
The rule-based trigger keeps running counters of repeated commands and errors, edit churn, distinct edited files, absence of state change, and repeated verification failures, which the judge's digest also reports. The judge is consulted only while the trigger is open, at most once every 5 turns; after each productive verdict, this interval grows by 5 turns, up to 20. A forced stop is checked before the judge and cannot be vetoed by it: once the trajectory's turns, counted across a redirect, reach $\kappa B_i$ with $\kappa=2.5$, and the strict stall rule holds, the rollout ends without a further redirect. A failure signature is the first error line of a failing tool output, such as a traceback, a failed test, or an assertion, with numbers, memory addresses, and file paths removed, so that two outputs that fail the same way match. The strict stall rule holds when three conditions hold at once: edit churn, with at least 4 edits to one file or edits to at least 3 files; dense error output, with at least 6 error outputs; and repetition, with at least 20\% of failing outputs repeating an earlier failure signature. The budget in \cref{eq:budget} uses a high percentile of passed turn counts rather than the maximum, so one unusually long success cannot inflate it, while repositories whose successful runs are typically long still receive a proportionally larger budget. The redirect is an LLM-authored summary of the agent's own actions that asks it to try a materially different hypothesis.

\section{Prompt Templates}
\label{appendix:prompts}

This appendix gives the prompts of the controller's model calls, along with the prefix that micro-skill injection adds to the agent's prompt. Slots filled at run time are shown in braces.

\subsection{Micro-skill Injection}

\paragraph{Proposer.} The proposer is called after task 4 and every 5 tasks thereafter, with the digests of the repository's 8 most recent tasks. The prompt's \texttt{fired} field is the need detector of \cref{sec:injection}.

\begin{PromptText}
Your task is to mine a durable, repository-specific set of micro-skills for a coding agent from digests of recently finished tasks in ONE repository. Each digest lists edited files, command templates, test invocations, and boolean flags, not raw logs.

Emit skills that a FUTURE task in this repository would want stated up front to avoid wasted commands: the canonical test invocation, whether a reproduction script is redundant, a required environment variable, a build or reinstall step that is needed (or not needed), a lint or format hook that reformats, etc. Do NOT emit task-specific fixes or which lines to change; emit only durable facts about the repository that hold across tasks.

For EACH skill return an object with:
  id           short kebab-case slug
  claim        one sentence stating what is true about this repository
  injection    the exact line to add to the agent's prompt, stated as a declarative fact
               about the repository (where things are, what the canonical command is).
               NEVER phrase it as a prohibition on exploration or tool use: do not tell
               the agent to avoid find/grep/searching/reading or to skip discovering files.
  high_stakes  true if a wrong belief would break the run (test command, build step)
  detector     {scope, confirm, contradict?, fired?}. scope is a regex over commands, or a
               built-in predicate (repro_written, importerror, badarg, ran_tests), that
               selects tasks where the skill is relevant; confirm is a regex whose match in
               commands and output means the skill held on this task; contradict (optional)
               means it was false; fired (optional) means the waste it prevents occurred.
  cite         at least 2 task IDs from the window on which confirm matches

Return at most 4 skills. Be conservative: a skill you cannot ground with a working detector and two citations is worse than no skill.

If a skill needs evidence that the digests do not contain, first define an extractor that surfaces it, {name, kind, pattern} with kind in {cmd_regex, out_regex, cmd_count}, and name it as the detector's scope or fired.

To revise an existing skill, re-emit it with the SAME id and change ONLY detector.contradict; it keeps its history. Any other change to an existing id is rejected.

Repository: @@{repository}@@
Existing skill IDs: @@{skill IDs}@@
Existing extractor names: @@{extractor names}@@
Task digests (chronological): @@{digests, as JSON}@@
Raw command excerpts that no current extractor covers: @@{commands}@@

Return a JSON object {"extractors": [...], "skills": [...]}.
\end{PromptText}

\noindent\begin{minipage}{\linewidth}
\paragraph{Micro-skill manual.} The manual is rendered before each task from the activated skills and placed ahead of the task prompt; \cref{appendix:skill-examples} shows individual skills.

\begin{PromptText}
# repository manual (@@{repository}@@), learned from prior tasks in this repository
- @@{skill 1}@@
- @@{skill 2}@@
...

@@{task prompt}@@
\end{PromptText}
\end{minipage}

\subsection{Tool Pruning}

\paragraph{Tool judge.} The tool judge is called before each task other than a re-audit on the unprotected, never-used tools that evidence has not yet masked; the task text is the benchmark's own task description, with links to the repository's pull requests, issues, commits, and files replaced by a placeholder. The agent receives the unmodified description, whose links are unreachable behind the answer-key guard (\cref{appendix:guard}).

\begin{PromptText}
You are configuring the toolset for an autonomous coding agent that is about to solve ONE specific issue by reading, editing, and testing source code in a single repository, headless and single-shot (no human, one sealed container, no sub-agents, scheduling, or external memory). Every loaded tool's schema is re-sent on every turn and costs tokens. Below are the issue and the candidate tools. For THIS issue, mark each tool:
  LOAD  plausibly useful to read, navigate, edit, run, or test code for this issue
  DROP  this issue will not need it
Be conservative: DROP only when you are confident the tool is irrelevant to this issue. Do NOT assume a tool is needed merely because it exists.

ISSUE: @@{task text, first 4,000 characters}@@
TOOLS: @@{one line per candidate tool: name and description}@@

Return ONLY a JSON object {tool_name: {"load": true|false, "reason": "<short>"}}.
\end{PromptText}

\subsection{Failure-loop Intervention}

Both calls read the same digest of the trajectory: its running counters, the last 24 actions, and, once $B_i$ is available, the trajectory's length relative to it. The digest records tool names, commands, search patterns, edited file names, and error signatures, but never file contents or the task text.

\paragraph{Trace judge.} The trace judge is called only while the rule-based trigger is open.

\begin{PromptText}
You are monitoring an AI coding agent that is partway through ONE task in a software repository. A rule-based trigger flagged its recent process as a possible unproductive loop. Read the digest of its recent actions and judge whether it is genuinely STUCK (repeating the same kind of edit, test, or failure with no progress toward passing tests, and unlikely to recover if it continues the same way) or still making real progress (changing failures, narrowing the problem, exploring new hypotheses).

Judge ONLY the workflow in the digest; do NOT infer the task's solution or which lines to change. Favor sparing the agent (stuck=false) when there is any sign of progress; return stuck=true with high confidence only for a clear, repetitive dead end.

Recent process digest:
turns=@@{t}@@ edit_churn=@@{c}@@ distinct_edit_files=@@{f}@@ error_outputs=@@{e}@@ test_runs=@@{r}@@
repeated_identical_failures=@@{k}@@ stagnation=@@{s}@@
Tasks in this repository usually pass within about @@{B_i}@@ turns; this trajectory is at @@{t}@@ turns, @@{t/B_i}@@x that length, and its tests are not passing yet. A trajectory far past the typical length with recurring failures is likely a dead-end loop, even if it touches many files.
Recent actions (oldest to newest): @@{last 24 actions}@@

Return ONLY a JSON object {"stuck": bool, "confidence": 0.0-1.0, "reason": "<short justification>"}.
\end{PromptText}

\paragraph{Redirect writer.} The redirect writer is called when the guard intervenes, to write a redirect targeted at the loop.

\begin{PromptText}
An AI coding agent is stuck in a repetitive, unproductive loop partway through one task in a software repository. You are given a digest of its recent actions only (tool names, shell commands, searched patterns, edited file names, and error signatures; never file contents or the task description). Write a SHORT redirect (2-4 sentences) addressed to the agent that (1) names the specific repetitive pattern you actually see in the digest (e.g., re-editing the same file, re-running the same test to the same failure) and (2) proposes a materially different next step (a different file or root cause to consider, re-reading the failing test's expectation, verifying a new hypothesis before editing again).

Do NOT tell the agent to stop, give up, or reduce effort; only redirect. Do NOT infer the task's solution, name specific lines to change, or reference any code or identifier you cannot see in the digest. Output ONLY the redirect as plain text.

Recent process digest: @@{digest, as for the trace judge}@@
\end{PromptText}

\section{Ablations of Meta-strategies}
\label{appendix:ablation}

Each single-meta-strategy variant uses the same parameters as the full stack (\cref{tab:config}), so no variant is tuned for its own ablation. \cref{tab:ablation-main} reports the cost and Pass@1 of each variant, paired against default execution over the same task--seed units as \cref{tab:main}. \cref{tab:ablation} reports how often each meta-strategy acted, and \cref{tab:ablation-units} defines its columns. All columns are per-stream means, but their units differ. \emph{Needed} counts events, one for each exercised task that needed a skill, and is therefore 33 to 37 times the number of confirmed skills beside it. \emph{Masked} is the number of tools still masked at the end of a stream, whereas \emph{Re-audits} counts tasks run with the full catalog, so the two are not comparable. Meta-strategies that a variant disables are reported as zero.

Within each setting, the counts of the combined stack remain close to those of the corresponding single-meta-strategy variant, which indicates that the three meta-strategies seldom interfere with one another's decisions. The counts also show how selectively each meta-strategy acts. About one in five proposed skills is retired within its stream, so the ledger revises its beliefs rather than accumulating them. The trace judge is consulted about 10 to 13 times per stream, but redirects follow on roughly one task in ten and aborts on fewer than one in twenty, since most escalations end in a productive or uncertain verdict.

\begin{table}[ht]
\centering
\caption{Meta-strategy activity of the ablation variants in the two Claude Code settings, as per-stream means. \cref{tab:ablation-units} defines the columns, and \cref{tab:ablation-main} gives the cost and Pass@1 of each variant.}
\label{tab:ablation}
\small
\setlength{\tabcolsep}{2.6pt}
\begin{tabular}{lrrrrrrrrr}
\toprule
 & \multicolumn{4}{c}{Micro-skills} & \multicolumn{2}{c}{Tool schemas} & \multicolumn{3}{c}{Failure loops} \\
\cmidrule(lr){2-5}\cmidrule(lr){6-7}\cmidrule(lr){8-10}
Method & Proposed & Confirmed & Needed & Retired & Masked & Re-audits & Judged & Redirects & Aborts \\
\midrule
\multicolumn{10}{l}{\emph{Claude Code, Sonnet 5}} \\
Micro-skills & 11.4 & 7.2 & 268 & 2.1 & 0.0 & 0.0 & 0.0 & 0.0 & 0.0 \\
Tool pruning & 0.0 & 0.0 & 0 & 0.0 & 15.1 & 3.3 & 0.0 & 0.0 & 0.0 \\
Failure-loop guard & 0.0 & 0.0 & 0 & 0.0 & 0.0 & 0.0 & 10.8 & 4.4 & 1.4 \\
Combined & 12.1 & 8.4 & 301 & 2.6 & 14.8 & 3.1 & 9.9 & 3.9 & 2.0 \\
\addlinespace
\multicolumn{10}{l}{\emph{Claude Code, Opus 4.8}} \\
Micro-skills & 9.6 & 6.5 & 214 & 1.7 & 0.0 & 0.0 & 0.0 & 0.0 & 0.0 \\
Tool pruning & 0.0 & 0.0 & 0 & 0.0 & 14.6 & 5.2 & 0.0 & 0.0 & 0.0 \\
Failure-loop guard & 0.0 & 0.0 & 0 & 0.0 & 0.0 & 0.0 & 13.1 & 4.8 & 1.7 \\
Combined & 10.3 & 7.1 & 241 & 2.2 & 14.5 & 4.3 & 12.2 & 5.3 & 2.2 \\
\bottomrule
\end{tabular}

\end{table}

\begin{table}[ht]
\centering
\caption{Column definitions for \cref{tab:ablation}. Every entry is a mean over the six repository streams of a variant.}
\label{tab:ablation-units}
\small
\begin{tabular}{@{}lllp{5.9cm}@{}}
\toprule
Group & Column & Unit & Definition \\
\midrule
Micro-skills & Proposed & skills / stream & candidate skills in the stream's final ledger \\
 & Confirmed & skills / stream & candidates that met the confirmation condition and were not retired \\
 & Needed & tasks / stream & exercised tasks that needed a skill, summed over skills \\
 & Retired & skills / stream & retired skills \\
\addlinespace
Tool schemas & Masked & schemas / stream & tools still masked at the last task, not a cumulative count \\
 & Re-audits & tasks / stream & tasks run as a full-catalog re-audit \\
\addlinespace
Failure loops & Judged & consultations / stream & escalations to the trace judge \\
 & Redirects & redirects / stream & redirects written into a trajectory \\
 & Aborts & terminations / stream & trajectories stopped before completion \\
\bottomrule
\end{tabular}
\end{table}

\raggedbottom
\section{Examples of Meta-strategy Decisions}
\label{appendix:skill-examples}

This appendix illustrates each meta-strategy with examples from SWE-Bench: micro-skills in (a)--(e), tool masks in (f) and (g), and failure-loop interventions in (h) and (i). Examples (a), (b), and (f)--(i) compare \method with default execution on the same task.

\subsection{Micro-skill Injection}

Each micro-skill example gives the skill as injected, its detectors as the proposer wrote them, and its lifecycle in the stream that learned it. Examples (a) and (b) are activated and injected, and each ends with one task that received the skill, compared with default execution. The remaining three illustrate the other outcomes: (c) is retired by a contradiction, (d) is confirmed but never clears the need gate, and (e) never meets the confirmation condition.

Tasks are numbered by their position in the stream, so the proposer runs after tasks 4, 9, 14, and so on. A detector is a regular expression over the task's commands and outputs unless it names a built-in predicate, such as {\fontencoding{T1}\selectfont\texttt{ran\_tests}}, or an extractor the proposer defined, whose pattern is given in parentheses; a dash marks a detector the proposer omitted. In the traces, each line that starts with \texttt{\$} or a tool name such as \texttt{Edit} is one tool call, indented grey text is its output, and a grey count in parentheses stands for tool calls that are not shown.

\medskip
\noindent\begin{minipage}{\linewidth}
\textbf{(a) sympy}\enspace Claude Code, Opus~4.8 \hfill\texttt{sympy-13615}
\blocklabel{\textbf{Skill}}
\begin{SkillText}
Besides pytest, tests can be run through the native runner `python -c "import sympy; sympy.test('sympy/.../test_x.py')"` or `python -m sympy.utilities.runtests`.
\end{SkillText}
\blocklabel{\textbf{Detectors}}
\begin{SkillText}
scope       ran_tests
confirm     sympy\.utilities\.runtests|sympy\.test\(
contradict  -
need        sympy_test_runner (sympy\.utilities\.runtests|sympy\.test\( in a command)
\end{SkillText}
\blocklabel{\textbf{Lifecycle}, 60 tasks}
\begin{SkillText}
proposed    after task 9, citing tasks 5 and 6
activated   on admission: confirmed on tasks 3, 5, and 6, whose edits do not overlap
            (matexpr.py | permutations.py | sparse.py); need 4/4
injected    tasks 10-60; the traces below are task 11
end         active; 10 confirmations and no contradiction over 10 exercised tasks
\end{SkillText}
\blocklabel{\textbf{Default}, 15 turns}
\begin{Trace}
  @@(10 tool calls: locate and edit Complement, run a reproduction)@@
$ python -m pytest sympy/sets/tests/test_sets.py -q 2>&1 | tail -15
  @@/opt/miniconda3/envs/testbed/bin/python: No module named pytest@@
$ python -m sympy.utilities.runtests sympy/sets/tests/test_sets.py 2>&1 | tail -20
  @@(deprecation warnings only; no tests run)@@
$ python -c "import sympy; sympy.test('sympy/sets/tests/test_sets.py')" 2>&1 | tail -20
  @@tests finished: 72 passed, 2 expected to fail, in 5.48 seconds@@
  @@(1 tool call: run the whole sympy/sets package)@@
\end{Trace}
\blocklabel{\textbf{\method}, 11 turns}
\begin{TraceTE}
  @@(9 tool calls: locate and edit Complement, run a reproduction)@@
$ python -c "import sympy; sympy.test('sympy/sets/tests/test_sets.py')" 2>&1 | tail -15
  @@tests finished: 72 passed, 2 expected to fail, in 5.37 seconds@@
\end{TraceTE}
\end{minipage}

\medskip
\noindent\begin{minipage}{\linewidth}
\textbf{(b) sphinx}\enspace Claude Code, Sonnet~5 \hfill\texttt{sphinx-7889}
\blocklabel{\textbf{Skill}}
\begin{SkillText}
The canonical way to reproduce an issue in this repo is to build a minimal Sphinx project under /tmp with a conf.py and run `python -m sphinx -b <builder> <srcdir> <outdir>`.
\end{SkillText}
\blocklabel{\textbf{Detectors}}
\begin{SkillText}
scope       python -m sphinx -b
confirm     python -m sphinx -b \S+ \S+ \S+
contradict  -
need        -
\end{SkillText}
\blocklabel{\textbf{Lifecycle}, 44 tasks}
\begin{SkillText}
proposed    after task 4, citing tasks 1-3
candidate   confirmed on tasks 1 and 3, but both edited a file named conf.py, so the two
            count once toward the 3 confirmations with non-overlapping edits
activated   after task 6, once tasks 5 (inheritance_diagram.py) and 6 (other.py, toctree.py)
            confirm it; need 4/4
injected    tasks 7-44; the traces below are task 15
end         active; 32 confirmations and no contradiction over 32 exercised tasks
\end{SkillText}
\blocklabel{\textbf{Default}, 11 turns}
\begin{Trace}
  @@(7 tool calls: read and edit mock.py, check git log and the installed version)@@
$ python -c "from sphinx.ext.autodoc.mock import _MockObject; ...; print(obj[T])"
  @@_MockObject.~T@@
  @@(1 tool call: a second inline check)@@
$ python -m pytest tests/test_ext_autodoc_mock.py -q 2>&1 | tail -20
\end{Trace}
\blocklabel{\textbf{\method}, 10 turns}
\begin{TraceTE}
  @@(5 tool calls: read mock.py, search git log, edit, run an inline check)@@
$ mkdir -p /tmp/genericmock/mymod; cat > /tmp/genericmock/conf.py <<'EOF' ...
    autodoc_mock_imports = ['missingdep'] ... EOF; cd /tmp/genericmock && python -m sphinx -b html . _build
  @@build succeeded, 1 warning.@@
$ git stash && cd /tmp/genericmock && python -m sphinx -b html . _build ...; git stash pop
  @@(2 tool calls: run tests/test_ext_autodoc_mock.py, clean up)@@
\end{TraceTE}
\end{minipage}

\medskip
\noindent\begin{minipage}{\linewidth}
\textbf{(c) sympy}\enspace Claude Code, Sonnet~5 \hfill retired after one injection
\blocklabel{\textbf{Skill}, high-stakes}
\begin{SkillText}
Run sympy tests with `python -m pytest sympy/<subpackage>/tests/test_<name>.py -q`; this is the canonical test invocation for this repo.
\end{SkillText}
\blocklabel{\textbf{Detectors}}
\begin{SkillText}
scope       pytest_module_used (python -m pytest sympy/.*test_.*\.py in a command)
confirm     python -m pytest sympy/.*test_.*\.py
contradict  ImportError|ModuleNotFoundError|no tests ran|error: unrecognized
need        -
\end{SkillText}
\blocklabel{\textbf{Lifecycle}, 60 tasks}
\begin{SkillText}
proposed    after task 4, citing tasks 2 and 4
confirmed   on admission: tasks 2, 3, and 4, whose edits do not overlap
            (function.py | matexpr.py, summations.py | permutations.py)
activated   after task 5, once the need gate has 4 exercised tasks
injected    task 6 only
retired     after task 6, whose trace contradicts it, which drives its support below zero
end         retired; 40 confirmations and 11 contradictions over 51 exercised tasks
\end{SkillText}
\blocklabel{\textbf{\method}, task 6, \texttt{sympy-13031}, 15 turns}
\begin{TraceTE}
  @@(11 tool calls: locate the stacking code, run reproductions, read git history)@@
$ python -m pytest sympy/matrices/tests/test_commonmatrix.py -q -k "hstack or vstack" 2>&1 | tail -20
  @@/opt/miniconda3/envs/testbed/bin/python: No module named pytest@@
$ source activate testbed 2>/dev/null; which python; python -c "import pytest" 2>&1; conda env list
  @@ModuleNotFoundError: No module named 'pytest'@@
  @@(1 tool call: run the tests with bin/test)@@
\end{TraceTE}
\end{minipage}
\par\smallskip
The instruction is wrong for this environment, which has no pytest: all 57 traces in the stream that invoke pytest fail with \texttt{No module named pytest}. The skill was nonetheless activated, because its confirming detector matches the pytest command itself rather than a passing result, so the agent's failed attempts on tasks 2--5 counted as confirmations. The first task to receive the skill tried the command and then probed for the module, and the probe's \texttt{ModuleNotFoundError} matched the contradicting detector. Weighted by $\lambda=5$, that single contradiction outweighs the decayed confirmations, and the skill left the manual after one injection. Example~(a) shows the native runner that another stream learned for the same repository.

\medskip
\noindent\begin{minipage}{\linewidth}
\textbf{(d) sympy}\enspace Claude Code, Opus~4.8 \hfill never clears the need gate
\blocklabel{\textbf{Skill}, high-stakes}
\begin{SkillText}
The sympy package is imported from the in-tree source, so edits to .py files take effect immediately; no `pip install`/`setup.py develop` reinstall is required before testing.
\end{SkillText}
\blocklabel{\textbf{Detectors}}
\begin{SkillText}
scope       ran_tests
confirm     python -m pytest\s+sympy/|sympy\.test\(
contradict  pip install|setup\.py\s+(develop|install)
need        reinstall_invoked (pip install|setup\.py\s+(develop|install) in a command)
\end{SkillText}
\blocklabel{\textbf{Lifecycle}, 60 tasks}
\begin{SkillText}
proposed    after task 9, citing tasks 3 and 6
confirmed   on admission: tasks 3, 5, and 6, whose edits do not overlap
            (matexpr.py | permutations.py | sparse.py); need 0/4
activated   never: no exercised task ever reinstalls, so the need rate stays 0 < 0.4
end         not activated; 10 confirmations, no contradiction, need 0/10
\end{SkillText}
\end{minipage}
\par\smallskip
This skill comes from the same stream as (a) and has the same confirmations. Its confirmations hold, but the agent never attempts the reinstall it warns against, so injecting it would add its tokens to every turn without preventing any waste.

\medskip
\noindent\begin{minipage}{\linewidth}
\textbf{(e) sphinx}\enspace Claude Code, Sonnet~5 \hfill never activated
\blocklabel{\textbf{Skill}, high-stakes}
\begin{SkillText}
The canonical test file for sphinx/domains/python.py is tests/test_domain_py.py; Python-domain edits are verified with `python -m pytest tests/test_domain_py.py`.
\end{SkillText}
\blocklabel{\textbf{Detectors}}
\begin{SkillText}
scope       domains/python\.py|test_domain_py
confirm     test_domain_py\.py
contradict  test_domain_py\.py.{0,20}(FAILED|ERROR|not found)
need        -
\end{SkillText}
\blocklabel{\textbf{Lifecycle}, 44 tasks}
\begin{SkillText}
proposed    after task 34, citing tasks 30 and 34
candidate   confirmed on tasks 30, 33, 34, 38, 39, 40, and 43, which edited
            python.py | docfields.py | python.py | __init__.py, inspect.py, python.py |
            python.py | python.py | python.py
end         candidate; 7 confirmations, only 2 with non-overlapping edits
\end{SkillText}
\end{minipage}
\par\smallskip
Every confirmation except one comes from a task that edited a file named \texttt{python.py}, so the evidence reflects one module that recurs across tasks rather than a practice that holds across the repository, and the skill is never activated.

\subsection{Tool Pruning}

Examples (f) and (g) show the tool mask over the first 4 tasks of the Django stream on Claude Code and Copilot CLI, together with the prompt tokens of each task's first request.

\medskip
\noindent\begin{minipage}{\linewidth}
\textbf{(f) django}\enspace Claude Code, Sonnet~5 \hfill first 4 tasks\quad tool pruning
\blocklabel{\textbf{Masked tools}}
\begin{SkillText}
AskUserQuestion, CronCreate, CronDelete, CronList, EnterPlanMode, EnterWorktree, ExitPlanMode, ExitWorktree, NotebookEdit, ScheduleWakeup, Skill, Task, TaskCreate, TaskGet, TaskList, TaskOutput, TaskStop, TaskUpdate, Workflow
\end{SkillText}
\blocklabel{\textbf{Default}, every task}
\begin{Trace}
  @@23 tools loaded; first request of each task: 42-46k prompt tokens@@
\end{Trace}
\blocklabel{\textbf{\method}, first-request prompt tokens (default in parentheses)}
\begin{TraceTE}
task 1  django-10097  23 tools loaded             @@42,866 (42,866)@@
task 2  django-10554  judge masks the 19 above    @@16,295 (43,351)@@
task 3  django-10880  judge masks the same 19     @@15,003 (42,059)@@
task 4  django-10914  evidence masks the same 19  @@15,299 (42,355)@@
                      (each unused for 3 tasks)
  @@loaded from task 2 on: Bash, Edit, Read, Write@@
\end{TraceTE}
\end{minipage}

\medskip
\noindent\begin{minipage}{\linewidth}
\textbf{(g) django}\enspace Copilot CLI, Sonnet~5 \hfill first 4 tasks\quad tool pruning
\blocklabel{\textbf{Masked tools}}
\begin{SkillText}
factories_manage, fetch_copilot_cli_documentation, list_agents, read_agent, run_factory, session_store_sql, skill, sql, task, write_agent
\end{SkillText}
\blocklabel{\textbf{Default}, every task}
\begin{Trace}
  @@19 tools loaded; first request of each task: 25-26k prompt tokens@@
\end{Trace}
\blocklabel{\textbf{\method}, first-request prompt tokens (default in parentheses)}
\begin{TraceTE}
task 1  django-10097  19 tools loaded             @@25,765 (25,760)@@
task 2  django-10554  judge masks the 10 above    @@11,347 (26,249)@@
task 3  django-10880  judge masks the same 10     @@10,066 (24,969)@@
task 4  django-10914  evidence masks the same 10  @@10,343 (25,248)@@
                      (each unused for 3 tasks)
  @@loaded from task 2: bash, create, edit, glob, grep, list_bash, read_bash, stop_bash, view@@
\end{TraceTE}
\end{minipage}

\subsection{Failure-loop Intervention}

Example (h) shows a redirect after which the agent tests its change and finishes in fewer turns than default execution, and (i) shows a forced stop on a task that default execution also fails.

\medskip
\noindent\begin{minipage}{\linewidth}
\textbf{(h) sphinx}\enspace Claude Code, Opus~4.8 \hfill\texttt{sphinx-11445}\quad redirect
\blocklabel{\textbf{Default}, 42 turns}
\begin{Trace}
  @@18 edits to sphinx/util/rst.py and 3 test runs before the task passes@@
\end{Trace}
\blocklabel{\textbf{\method}, 31 turns}
\begin{TraceTE}
  @@turns 1-18: 11 edits to sphinx/util/rst.py, no test run, the same failure repeated 3 times@@
Redirect (trace judge: stuck, confidence 0.72)
  @@Process check (automated): so far you've made 11 edits across 1 file(s) ... and run 0@@
  @@verification(s) ... try a MATERIALLY DIFFERENT hypothesis: re-read the failing test's expectation ...@@
$ grep -rn "prepend_prolog\|rst_prolog\|docinfo" /testbed/tests/test_util_rst.py
$ python -c "from sphinx.util.rst import prepend_prolog; ..."
$ python -m pytest tests/test_util_rst.py -q 2>&1 | tail -15
  @@7 passed in 0.65s@@
\end{TraceTE}
\end{minipage}

\medskip
\noindent\begin{minipage}{\linewidth}
\textbf{(i) xarray}\enspace Claude Code, Opus~4.8 \hfill\texttt{xarray-6798}\quad forced stop
\blocklabel{\textbf{Default}, 32 turns, fails}
\begin{Trace}
  @@6 edits to xarray/core/coordinates.py and 3 test runs; the task still fails@@
\end{Trace}
\blocklabel{\textbf{\method}, 25 turns, fails}
\begin{TraceTE}
  @@learned turn budget B_i = 10.25 turns; forced stop at 2.5 B_i@@
  @@by then: 7 edits to xarray/core/coordinates.py, 14 error outputs, the same failure repeated 3 times@@
Stop: past 2.5 B_i, and the strict stall rule fires
  @@the task fails, as it does under default execution, 7 turns sooner@@
\end{TraceTE}
\end{minipage}

\section{Harness-specific Customization}
\label{appendix:codex}

\method reaches each agent only through interfaces its harness already exposes: the task prompt, the tool configuration applied at launch, and the event stream the harness emits while it runs. The three meta-strategies and normalized trace format are shared across harnesses, and harness-specific adapters implement the available controls. Every paid controller call uses the same model and routing as the task agent.

Vendor-specific event streams are converted into a common representation containing tool names, arguments, observations, model outputs, usage, and grading outcomes. Meta-strategy implementations consume only this normalized interface. All experiments use Claude Code 2.1.170, Codex 0.153.1, and GitHub Copilot CLI 1.0.84-7.

\paragraph{Answer-key guard.}
\label{appendix:guard}
Web-Bench grading specifications and SWE-Bench fix commits are public on GitHub, so an agent with network access could read the answer instead of solving the task. On both benchmarks, every method, including default execution, therefore runs behind the same guard on all three harnesses: the task container resolves GitHub hosts to an unreachable address, so any command that looks them up by name fails to connect. Claude Code and Copilot CLI additionally run a pre-execution hook that refuses shell commands referencing GitHub; Codex~0.153.1 accepts a hook configuration but did not invoke it in our tests, likely a limitation of this version, so it relies on the first layer alone. 

\paragraph{Claude Code.}
Claude Code runs headless and emits a JSON event stream, which the controller reads live and keeps as the task's trace. The tool catalog comes from the stream's session-start event.
\begin{itemize}[noitemsep,topsep=2pt,leftmargin=5mm]
\item \emph{Micro-skills.} The rendered manual is prepended to the task prompt.
\item \emph{Tool pruning.} Masked tools are passed to the CLI's disallowed-tools option, which removes their schemas from every request.
\item \emph{Failure-loop intervention.} The guard reads the stream as it arrives. To redirect, it stops the CLI and starts a new invocation in the same container, with the redirect in its prompt and the agent's edits still on disk. A second stuck verdict ends the task. Token usage from both invocations is summed.
\end{itemize}

\paragraph{Codex.}
Codex also runs headless with a JSON event stream. It differs from the other two harnesses in one interface: its tools cannot be removed one at a time, because groups of related tools and instructions can only be switched on or off in its configuration file.
\begin{itemize}[noitemsep,topsep=2pt,leftmargin=5mm]
\item \emph{Micro-skills.} The rendered manual is prepended to the task prompt, as on Claude Code.
\item \emph{Tool pruning.} The pruning unit is a capability package rather than an individual tool: sub-agent collaboration, goal tracking, image input, bundled skills, and the base prompt's guidance for interactive chat, which a headless run never uses. The rules of \cref{sec:untooling} apply to packages unchanged. The mask is written into each task's configuration file.
\item \emph{Failure-loop intervention.} As on Claude Code, a redirect starts a new invocation in the same container, and a second stuck verdict ends the task.
\end{itemize}

\paragraph{Copilot CLI.}
Copilot CLI's event stream does not report token usage, so the controller reads usage and the tool catalog from the CLI's debug log.
\begin{itemize}[noitemsep,topsep=2pt,leftmargin=5mm]
\item \emph{Micro-skills.} The rendered manual is prepended to the task prompt, as on Claude Code.
\item \emph{Tool pruning.} Masked tools are passed to the CLI's excluded-tools option, which removes their schemas from the request; its deny-tool option would only refuse execution while still billing each schema on every turn.
\item \emph{Failure-loop intervention.} As on Claude Code, a redirect starts a new invocation in the same container, and a second stuck verdict ends the task.
\end{itemize}

\end{document}